\documentclass{article}

\usepackage{microtype}
\usepackage{latexsym}
\usepackage{balance}
\usepackage{enumitem}
\usepackage{wrapfig}
\usepackage{lineno}

\usepackage{amsmath}
\usepackage{amssymb}
\usepackage{amsfonts}
\usepackage{amsthm}
\usepackage{mathtools}
\usepackage{bm}
\usepackage{dsfont}

\usepackage{graphicx}
\usepackage{subcaption}
\usepackage{tikz}
\usetikzlibrary{positioning}
\usepackage{pgf}
\usepackage{xfp}

\usepackage{booktabs}
\usepackage{tabularx}
\usepackage{multirow}
\usepackage{makecell}
\usepackage{array}
\usepackage{colortbl}

\usepackage{tcolorbox}
\tcbuselibrary{breakable}
\tcbset{
  colback=red!15,
  boxrule=0pt,
  arc=0pt,
  left=2pt,
  right=2pt,
  top=1pt,
  bottom=1pt
}

\usepackage{xspace}
\usepackage{soul}
\usepackage[normalem]{ulem}
\usepackage{pifont}      
\usepackage{wasysym}
\usepackage{resizegather}
\usepackage{lipsum}

\usepackage{url}

\usepackage{hyperref}
\usepackage{hyperxmp}
\usepackage[capitalize,noabbrev]{cleveref}

\usepackage[textsize=tiny]{todonotes}

\PassOptionsToPackage{table,xcdraw}{xcolor}

\usepackage[accepted]{icml2026}

\tcbset{colback=red!15, boxrule=0pt, arc=0pt, left=2pt, right=2pt, top=1pt, bottom=1pt} 
\newcolumntype{X}{>{\centering\arraybackslash}m{0.07\linewidth}}
\definecolor{darkgreen}{HTML}{006400}

\newcommand{\proposed}{\textsc{Persona2Web}}
\newcommand{\pweb}{\mathcal{P}_{\text{web}}}
\newcommand{\ppref}{\mathcal{P}_{\text{pref}}}
\newcommand{\pavg}{\mathcal{P}_{\text{avg}}}

\theoremstyle{plain}

\theoremstyle{definition}

\theoremstyle{remark}

\icmltitlerunning{Persona2Web: Benchmarking Personalized Web Agents for Contextual~Reasoning with User History}

\begin{document}

\twocolumn[
  \icmltitle{Persona2Web: Benchmarking Personalized Web Agents for Contextual~Reasoning with User History}



  \icmlsetsymbol{equal}{*}

  \begin{icmlauthorlist}
    \icmlauthor{Serin Kim}{yyy}
    \icmlauthor{Sangam Lee}{yyy}
    \icmlauthor{Dongha Lee}{yyy}
  \end{icmlauthorlist}

  \icmlaffiliation{yyy}{Department of Artificial Intelligence, Yonsei University, Seoul, Republic of Korea}
  \icmlcorrespondingauthor{Dongha Lee}{donalee@yonsei.ac.kr}

  \icmlkeywords{Web Agent, Personalized Web Agent, Personalization, Benchmark}

  \vskip 0.3in
]
\newcommand{\bespoke}{\textsc{Bespoke}\xspace}
\newcommand{\fix}{\marginpar{FIX}}
\newcommand{\new}{\marginpar{NEW}}
\newcommand{\cmark}{\textcolor{green!60!black}{\ding{51}}} 
\newcommand{\xmark}{\textcolor{red}{\ding{55}}}   


\printAffiliationsAndNotice{}  

\begin{abstract}
Large language models have advanced web agents, yet current agents lack personalization capabilities.
Since users rarely specify every detail of their intent, practical web agents must be able to interpret ambiguous queries by inferring user preferences and contexts.
To address this challenge, we present \proposed, the first benchmark for evaluating personalized web agents on the real open web, built upon the \textit{clarify-to-personalize} principle, which requires agents to resolve ambiguity based on user history rather than relying on explicit instructions.
\proposed\ consists of:
(1) user histories that reveal preferences implicitly over long time spans,
(2) ambiguous queries that require agents to infer implicit user preferences, and
(3) a reasoning-aware evaluation framework that enables fine-grained assessment of personalization.
We conduct extensive experiments across various agent architectures, backbone models, history access schemes, and queries with varying ambiguity levels, revealing key challenges in personalized web agent behavior.
For reproducibility, our codes and datasets are publicly available at \href{https://serin-kimm.github.io/Persona2Web/}{[CODE]}.
\end{abstract}

\section{Introduction}
\label{sec:introduction}
\begin{figure}[!tb]
    \centering
    \includegraphics[width=\columnwidth]{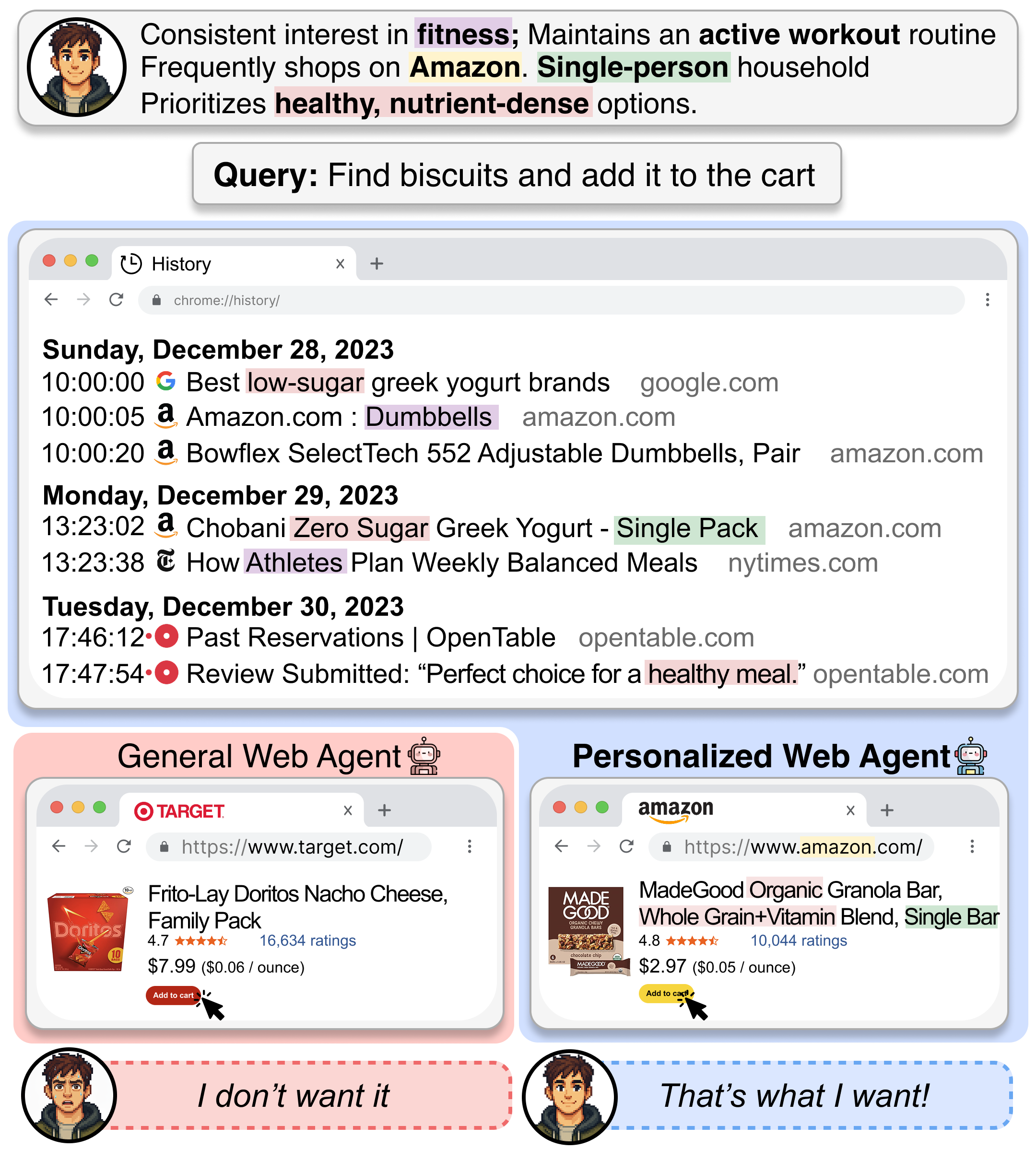}
    \caption{A personalized web agent generates user-specific responses by leveraging user history, whereas a general web agent often produces generic or random outputs that fail to align with user preference.}
    \label{fig:motivation}
\end{figure}
Large Language models (LLMs) have progressed beyond imitating human reasoning and are beginning to act autonomously on behalf of humans in complex, real-world tasks~\cite{xi2023risepotentiallargelanguage}. 
A representative paradigm of such advancement is the web agent, an autonomous system designed to execute multi-step workflows on web environment that require human intervention~\cite{Wang_2024}. 
At the core of these capabilities, LLMs enable the agent to interpret web contents~\cite{Gur2022UnderstandingHW, gur2024realworldwebagentplanninglong} and understand task-specific instructions~\cite{ouyang2022traininglanguagemodelsfollow}, which enhances the robustness of web agents.

However, these advances remain insufficient for fully practical use in real-world scenarios. 
In real-world scenarios, users rarely specify every detail of their intent, assuming that systems understand their implicit context~\cite{Kelly2003ImplicitFF, Hu2008CollaborativeFF, 10.1145/2736277.2741646, Radlinski2017ATF, 10.1145/3544548.3581388, deng-etal-2024-multi}.
To be genuinely practical, web agents must be able to accurately interpret even such ambiguous queries by inferring the user context and preferences.
Yet, they often provide generic and sometimes arbitrary responses to ambiguous queries~\cite{Song2007IdentifyingAQ, Qu2018AnalyzingAC}. 
In Figure~\ref{fig:motivation}, when given the ambiguous query, the general web agent (Left) fails to clarify the query and selects a random high-sugar snack that has no relation to the user's preferences. 
Such behavior yields unsatisfactory outcomes for the user with strong preference for health-conscious options. 
Personalization can directly address this limitation by enabling agents to leverage user history when resolving ambiguity.
When given the same query, the personalized web agent (Right) leverages the user history and selects snacks that align with the user’s health-conscious preferences.

Despite the importance of personalization in web agents, existing benchmarks fail to provide a pertinent evaluation framework for these capabilities. 
Specifically, they suffer from two fundamental limitations.
First, the majority of existing web agent benchmarks do not provide the user context that reveals user's preference or behavioral patterns which are essential for agents to personalize their responses. 
Even when some form of context is provided~\cite{Zhou2023WebArenaAR, Yao2024benchAB}, it does not reflect realistic user behavior that agents would encounter in practice.
Second, current benchmarks~\cite{Deng2023Mind2WebTA, He2024WebVoyagerBA} rely on fully specified queries that overlook the ambiguity inherent in real user-agent interactions.
This eliminates the need for agents to clarify implicit queries based on user context, which is the core challenge of personalization.
Consequently, these benchmarks cannot distinguish between agents that genuinely understand and apply user preferences and those that simply follow the instructions.

Motivated by these limitations, we introduce the first personalized web agent benchmark, \proposed, specifically designed for evaluating personalization capability of the web agent on the real open web. 
Central to this benchmark is the principle of \textit{clarify-to-personalize}.
This elicits personalization by challenging agents to interpret ambiguous queries and fill in missing details based on user history, rather than simply executing fully specified instructions.
Specifically, \proposed\ consists of three core components.
\textbf{(1) User History:}
Our rigorously-constructed user history extends simple browsing logs into richer and more detailed contexts. 
It reveals user preferences in implicit and indirect ways over long time spans, rather than providing them explicitly at once.
\textbf{(2) Ambiguous Query:}
Our benchmark intentionally conceals specific parts of the query to let the agent clarify based on user history.
This design assesses whether the agent can resolve ambiguity without forcing the users to provide exhaustive details.
\textbf{(3) Reasoning-aware Evaluation:}
Beyond simple task completion, our evaluation framework comprehensively examines reasoning traces through structured rubrics to properly distinguish personalization failures from navigation failures.
This enables fine-grained assessment of personalization in web agents.

Evaluating both general and personalized web agents on \proposed\ reveals a fundamental gap in current personalization capabilities.
Without access to user history, all agents fail completely on ambiguous queries, achieving a 0\% success rate, which confirms that agents cannot resolve ambiguity without user context. 
Even when user history is provided, performance improves only marginally, reaching just a 13\% success rate at best.
These observations indicate that simply supplying user history is insufficient and highlight the need for methods that enable agents to effectively leverage user context for personalization.
Furthermore, task completion alone cannot capture personalization capability, as agents with similar success rates excel at personalization but fail at navigation, or the reverse.
This underscores the necessity of our reasoning-aware evaluation framework for personalized web agents.

Our key contributions are as follows:
\begin{itemize}[leftmargin=*,topsep=2pt,itemsep=2pt,parsep=0pt]

    \item {We propose \proposed, the first benchmark for evaluating personalized web agents on the real open web, comprising user histories, personalization query sets, and evaluation framework.}

    \item {Our queries follow \textit{clarify-to-personalize} principle that deliberately embeds ambiguity, requiring the agent to infer implicit contexts from user history without explicit cues.}
    
    \item {We conduct extensive experiments across diverse backbone models, history access schemes, and query ambiguity levels to identify key challenges in personalized web agent behavior.}
\end{itemize}

\section{Related work}
\label{sec:relatedwork}
\begin{table}[t]
\caption{Comparison between \proposed\ and existing web agent benchmarks. \proposed\ offers extensive domain and website coverage, enables evaluation on the open web, and supports personalization assessment. Parenthesized numbers next to the environment indicate the number of websites used; simulated, function-call, and dialogue agent therefore omit website counts. Domain counts are recalculated based on our own criteria.}
\label{tab:benchmarks_2}
\centering
\small
\setlength{\tabcolsep}{3pt}
\renewcommand{\arraystretch}{1.05}

\begin{tabular}{%
>{\raggedright\arraybackslash}p{0.27\columnwidth}
>{\centering\arraybackslash}m{0.13\columnwidth}
>{\centering\arraybackslash}m{0.26\columnwidth}
>{\centering\arraybackslash}m{0.22\columnwidth}}
\toprule
\textbf{Benchmark} 
& \textbf{Domains}
& \textbf{Environment} 
& \textbf{Personalization} \\
\midrule
WebLINX
& -- & Cached (155) & \color{red}\ding{55} \\

Mind2Web
& 31 & Cached (137) & \color{red}\ding{55} \\

WebShop
& 1 & Simulated & \color{red}\ding{55} \\

WebArena
& 6 & Simulated & \color{red}\ding{55} \\

WebVoyager
& 5 & Open web (15) & \color{red}\ding{55} \\

AssistantBench
& 15+ & Open web (258) & \color{red}\ding{55} \\

WebCanvas
& 19 & Open web (69) & \color{red}\ding{55} \\

PersonalWAB
& 1 & Function-call & \color{darkgreen}\ding{51} \\

Apollonion
& 6 & Dialogue agent & \color{darkgreen}\ding{51} \\
\midrule
\textbf{\proposed}
& \textbf{21} & \textbf{Open web (105)} & \color{darkgreen}\ding{51} \\
\bottomrule
\end{tabular}
\end{table}

\begin{figure*}[t]
    \centering
    \includegraphics[width=\textwidth]{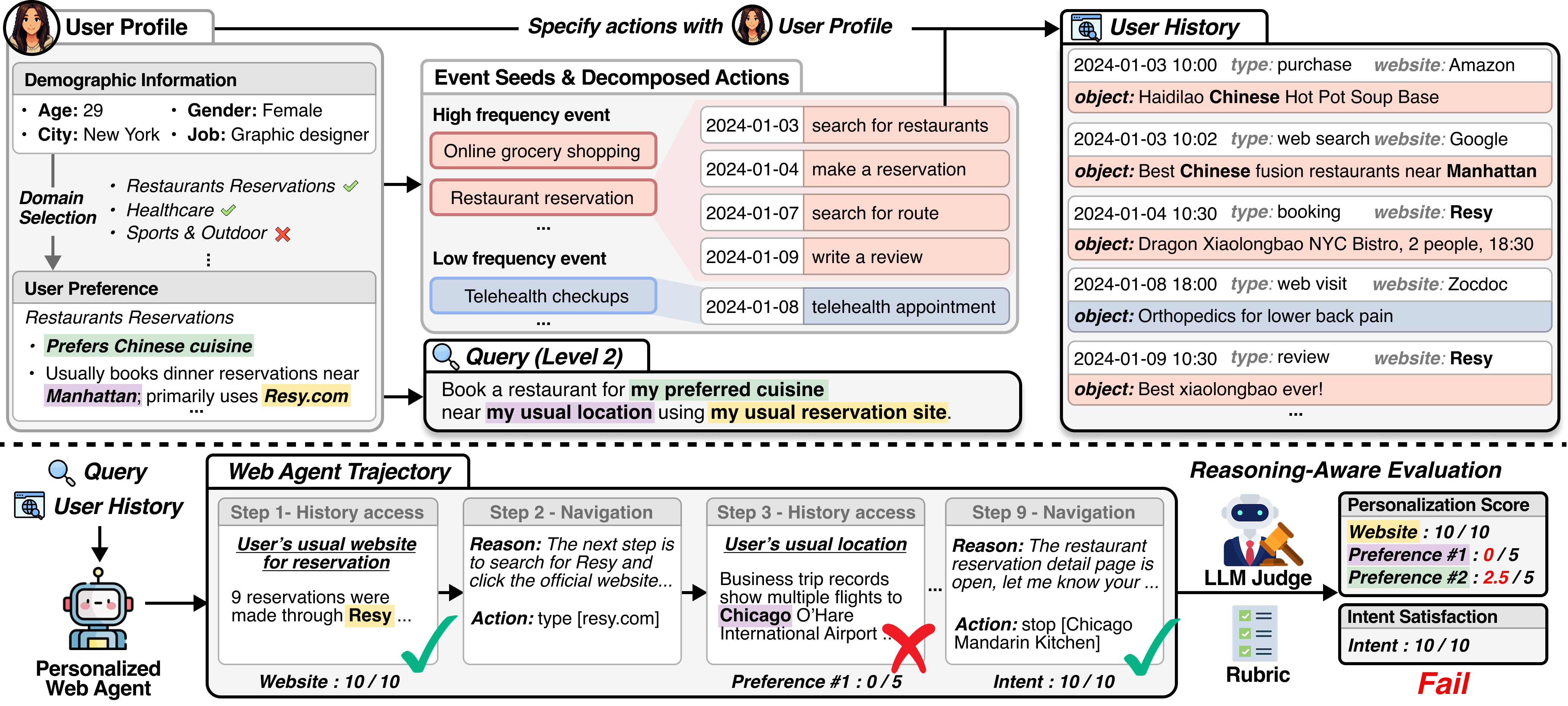}
    \caption{Overview of \proposed~construction pipeline and reasoning-aware evaluation process.}
    \label{fig:datasetconstruction}
\end{figure*}
\paragraph{Web agent benchmark.}

Evaluating the web agents presents inherent challenges due to the dynamic nature of the web.
To mitigate this issue, prior benchmarks have primarily relied on simulated or cached environments (Table~\ref{tab:benchmarks_2}).
MiniWOB++~\cite{Shi2017WorldOB, DBLP:journals/corr/abs-1802-08802} provides simplified HTML-based tasks to test interaction skills.
WebArena~\cite{Zhou2023WebArenaAR} introduces a self-hosted simulated platform with end-to-end executable tasks.
Mind2Web~\cite{Deng2023Mind2WebTA} caches real-world webpages and provides realistic tasks demanding multi-step execution.
Recent studies have shifted toward benchmarks on the real open web, introducing corresponding evaluation methods to handle its dynamic nature.
WebCanvas~\cite{Pan2024WebCanvasBW} defines mandatory key nodes and evaluates performance at each node, while WebVoyager~\cite{He2024WebVoyagerBA} integrates visual information into observation space and evaluates through LLM judge.

Beyond the environment limitations, existing evaluation methodologies fail to capture the complexity of web agent behavior.
Current web agent benchmarks assess performance based on step-wise actions~\cite{Gou2025Mind2Web2E} or the final outcomes~\cite{Yao2022WebShopTS, Zhou2023WebArenaAR, He2024WebVoyagerBA} making it impossible to discern whether failures stem from navigation errors, personalization deficiencies, or other reasoning breakdowns.
Unlike closed settings, open-web navigation often admits multiple trajectories that lead to correct outcomes, making action-wise or outcome-based evaluation inappropriate.
\vspace{-1em}
\paragraph{Personalization in LLMs.}
Personalization has been explored in memory-augmented LLMs~\cite{Kumar2024LongLaMPAB, Wu2024LongMemEvalBC, Pan2025OnMC, Huet2025EpisodicMG} and dialogue agents~\cite{Zhang2018PersonalizingDA, Xu2022LongTN, Lee2023PromptedLA, Chen2024ApollonionPD, Zhao2025DoLR}, typically defined as the ability to retrieve and reason over user-relevant information.
LaMP~\cite{Salemi2023LaMPWL} introduces a benchmark with user profiles across multiple domains, and LongLaMP~\cite{Kumar2024LongLaMPAB} extends it with a focus on the text generation tasks.
In dialogue settings, LongMemEval~\cite{Wu2024LongMemEvalBC} evaluates memory abilities based on chat history, Apollonion~\cite{Chen2024ApollonionPD} measures personalization through embedding-based similarity with user profiles, and PrefEval~\cite{Zhao2025DoLR} requires models to infer implicitly expressed preferences.
PersonalWAB~\cite{Cai2024LargeLM} explores personalization based on the user's history, but focuses on abstract function calls rather than web interface interactions.
Yet none of these benchmarks address web navigation scenarios, where user preferences are inferred from interaction patterns and apply them in the action planning and execution.
Motivated by these limitations, our work moves beyond response generation to encompass realistic web navigation tasks grounded in user context.

\section{\proposed}
\label{sec:persona2web}
\begin{table*}[t]
\caption{{Examples of a \proposed\ query set.
Each task provides 3 queries with varied ambiguity levels. Web. and Pref. denote website and preferences, respectively. Level 0 is a clear query that includes all explicit cues.}}
\label{tab:queryexample}
\centering
\resizebox{1.02\linewidth}{!}{
\begin{tabular}{cccl}
\toprule
\textbf{Level} & \textbf{Web.} & \textbf{Pref.} & \multicolumn{1}{c}{\textbf{Example Query}} \\
\midrule
0    
  & \ding{51} & \ding{51} & At \textbf{zocdoc.com}, search for doctors \textbf{near Southside Jacksonville} who is a \textbf{female}. \\
1
  & \ding{55}  & \ding{51} & Search for doctors \textbf{near Southside Jacksonville} who is a \textbf{female} at my preferred website.\\
2
  & \ding{55}  & \ding{55}  & Search for doctors in \textbf{my usual area} that match \textbf{my preferred provider gender} at my preferred website. \\
\bottomrule
\end{tabular}
}
\end{table*}
In this section, we present \proposed, a benchmark designed to evaluate personalized web agents under realistic, open-web environment.
It comprises three core components: user history, ambiguous query, and reasoning-aware evaluation.
Figure~\ref{fig:datasetconstruction} illustrates the overall pipeline.

\subsection{Realistic User History}
\label{subsec:userhistory}
We construct user context in the form of web browsing history, as this format most closely reflects how real web agents would access user information in practice.
While dialogue-based benchmarks~\cite{kwak2025toolhaystackstresstestingtoolaugmentedlanguage} incorporate user context through explicit verbal cues, browsing history encodes such information implicitly with web behaviors.
This requires agents to infer user preferences from actions rather than relying on explicit linguistic cues.
To this end, we construct user history through a multi-stage generation pipeline (see Tables~\ref{tab:domainpref}, \ref{tab:eventseed}, \ref{tab:actiondecomp}, and \ref{tab:memorybank} for detailed prompts).
The key principle is that user preferences should be \textit{revealed implicitly} from contextual cues distributed across the history, rather than stated explicitly.

\paragraph{User profile generation.}
The first stage constructs a user profile consisting of \textit{demographic information} $\mathrm{Dem}(u)$ and \textit{domain preferences} $\mathrm{Pref}(u)$.
Demographic attributes are evenly distributed to ensure variability across users.
Domain preference represents the user’s domain-specific preferences and patterns.
We define 21 domains reflecting typical web user experiences.

Given $\mathrm{Dem}(u)$ and the predefined domain set $\mathcal{D} = \{d_1, d_2, \dots, d_{21}\}$, the LLM $\mathcal{M}$ selects $K$ relevant domains with rationales\footnote{We use GPT-4o as $\mathcal{M}$ throughout the data generation pipeline.}:
\begin{equation*}
\{ (d^{(k)}, \rho^{(k)}) \}_{k=1}^{K} = \mathcal{M}\big(\mathrm{Dem}(u), \mathcal{D}\big),
\end{equation*}
where $d^{(k)}$ denotes $k$-th selected domain and $\rho^{(k)}$ is the rationale describing its relevance.
$\mathcal{M}$ then generates domain-specific preferences using domain prompt ${\pi^{(k)}}$ that specifies what attributes to include in domain $k$'s preferences:
\begin{equation*}
\mathrm{Pref}(u) = \left\{ \mathcal{M}(d^{(k)}, \pi^{(k)}, \rho^{(k)}, \mathrm{Dem}(u)) \right\}_{k=1}^{K}
\end{equation*}
The user profile is $\mathrm{Profile}(u) = \big(\mathrm{Dem}(u), \mathrm{Pref}(u)\big)$ as a result.
In total, we construct 50 distinct user profiles.

\paragraph{Event seeds and action decomposition.}
Since preferences are reflected in event occurrence, randomly generating events without event seed or directly importing external data~\cite{sap-etal-2020-recollection} fails to capture consistent preference.
We introduce event seeds that define recurring activity patterns based on user preferences and routines.
We categorize events into high-frequency (daily activities like commuting, shopping) and low-frequency types (irregular events like travel, medical checkups).

Given the $\mathrm{Dem}(u)$ and the set of rationales within all selected domains $\{\rho^{(k)}\}_{k=1}^K$, LLM generates two event sets: the high-frequency events $\mathcal{E}^{\mathrm{HF}}_u$ and the low-frequency events $\mathcal{E}^{\mathrm{LF}}_u$, as follows.
\begin{equation}
    (\mathcal{E}^{\mathrm{HF}}_u, \; \mathcal{E}^{\mathrm{LF}}_u) 
= \mathcal{M} \big(\mathrm{Dem}(u), \{\rho^{(k)}\}_{k=1}^K \big)
\end{equation}
We selectively apply cancellations and modifications to approximately 10\% of the entire record to reflect the noise and inconsistencies present in real-world user histories.
Each event $E_i$ is decomposed into a sequence of actions:
\begin{equation*}
(a_{i,1}, a_{i,2}, \dots, a_{i,L_i})
\;=\;
\mathcal{M}\big(E_i),
\quad i = 1, \dots, N
\end{equation*}
where $a_{i,j}$ denotes the $j$-th action derived from event $E_i$, and $L_i$ is the number of actions in the event $E_i$.
Actions from each event are dispersed across different temporal points over a year.
This prevents related actions from being clustered together and instead requires agents to integrate remote histories to recognize recurring patterns.

\paragraph{User history generation.}
We structure user history with four elements as shown in Figure~\ref{fig:datasetconstruction}: timestamp, type, object, and website.
The \textit{type} field includes five categories, web search, web visit, purchase, booking, and review and rating, capturing a broad spectrum of actions commonly observed in web environments.
\textit{Object} provides the detailed information of the target entity for each action type.

\subsection{Ambiguous Query Sets}
\label{subsec:querysets}
A key distinction of our work lies in intentionally masking explicit values, as the ambiguity of a query should be resolved through the user’s history.
Based on this \textit{clarify-to-personalize} approach, we first generate a query with all explicit cues, then create variants by adjusting the level of ambiguity.
Each query set consists of three levels (Table~\ref{tab:queryexample}): \textbf{Level 0} (clear query), where both website and preference constraints are explicit; \textbf{Level 1}, where preference constraints are explicit while website is masked; and \textbf{Level 2}, where both are masked.
Level 2 is the target query on which personalized web agents should perform well.

As shown in Tables~\ref{tab:stat_all} and~\ref{tab:domain_coverage}, \proposed\ provides an extensive, carefully constructed dataset for evaluating personalized web agents across diverse user activities and domains.
Notably, only 3.56\% of history entries contain preference values as exact string matches.
This low ratio attests that user preferences are embedded implicitly through contextual cues rather than as explicit textual signals.
Additional validation of user profiles and user histories is conducted in Appendix~\ref{appendix:datasetvalidation}.
The results show that the generated user histories exhibit broad topical coverage, and rich variation.
Human evaluation (Table~\ref{tab:dataset_realism}) further demonstrates that our multi-stage generation pipeline improves the realism of generated histories over naive generation.

\begin{table}[t]
\centering
\caption{Overall statistics of \proposed. Histories/user, subdomains/user, and websites/user denote averages across user. Domains are reported as domain/subdomain counts. Exact match ratio denotes the average proportion of history entries in which a preference value appears by exact string match.}
\label{tab:stat_all}
\small
\begin{tabular*}{\columnwidth}{@{\extracolsep{\fill}}lclc}
\toprule
\textbf{Statistic} & \textbf{Value} & \textbf{Statistic} & \textbf{Value} \\
\midrule
Users & 50 
& Queries & 150 \\
History entries & 102,568 
& Histories / user & 2,051 \\
Domains & 7 / 21 
& Subdomains / user & 17 \\
Websites / user & 35 
& Exact match ratio & 3.56\% \\
\bottomrule
\end{tabular*}
\end{table}
\begin{table}[t]
\centering
\caption{Domain coverage of user histories and profiles. Mean values indicate average of history entries and profiles across users.}
\label{tab:domain_coverage}
\small
\begin{tabular*}{\columnwidth}{@{\extracolsep{\fill}}lrrrr}
\toprule
\multirow{2.5}{*}{\textbf{Domain}} 
& \multicolumn{2}{c}{\textbf{User history}} 
& \multicolumn{2}{c}{\textbf{User profile}} \\
\cmidrule(lr){2-3} \cmidrule(lr){4-5}
& \textbf{Mean} 
& \textbf{Total} 
& \textbf{Mean} 
& \textbf{Total} \\
\midrule
E-commerce & 345 & 17,272 & 4 & 179 \\
Travel/Transportation & 659 & 32,943 & 4 & 201 \\
Weather/Maps & 271 & 13,546 & 1 & 50 \\
Health/Medical & 113 & 5,670 & 1 & 50 \\
Entertainment/Media & 261 & 13,069 & 3 & 150 \\
Education/Academia & 116 & 5,816 & 1 & 49 \\
Search/Community & 299 & 14,950 & 2 & 98 \\
\bottomrule
\end{tabular*}
\end{table}

\subsection{Reasoning-aware Evaluation}
\label{subsec:evaluation}
Accurate evaluation of personalized web agents requires distinguishing personalization failures from navigation failures.
However, since task completion depends on both personalization and navigation, errors in either are compounded in the final outcome, making the distinction impossible from task success alone.
Our evaluation framework examines the full trajectory including reasoning traces based on three metrics: \textbf{personalization score}, \textbf{intent satisfaction}, and \textbf{success rate}.
For each metric, we employ scoring rubrics that assign discrete point levels based on whether each criterion is met.
To apply these rubrics, we use GPT-5-mini as an LLM judge that receives the agent's full trajectory along with the rubrics and outputs scores for each metric.
Detailed rubrics are provided in Appendix~\ref{appendix:rubric}.

\paragraph{Personalization scores.}
To evaluate personalization, we define two Personalization Scores (PS), $\pweb$ and $\ppref$.
$\pweb$ determines whether the agent recognizes websites that align with recurring usage patterns based on the user history.
$\ppref$ further examines whether the agent identifies items or attributes that best reflect user preferences.
For websites requiring location-aware navigation (e.g., Amazon, Resy), we additionally evaluate whether the agent correctly personalizes based on user location from history, even without an explicit query cue.
Since personalization involves multiple steps to retrieve relevant history and utilize it, which can fail independently, each score is computed using two rubrics: \textit{retrieval accuracy}, which assesses whether the agent accesses the correct histories, and \textit{utilization accuracy}, which evaluates whether the retrieved information is appropriately incorporated into the navigation process. 
These rubrics require reasoning traces to identify whether failures occur during retrieval or utilization.
\vspace{-1em}
\paragraph{Intent satisfaction.}
Intent satisfaction measures task accuracy apart from personalization.
Considering the open-web environment, agents may formulate correct plans but fail due to external factors (website unavailability, dynamic content).
The rubric assigns full credit for complete task success and partial credit when the agent attempts the correct action but is blocked by external factors.
\vspace{-1em}
\paragraph{Success rate.}
Success Rate (SR) counts cases with full scores in both PS and IS.
The task is deemed successful only when the agent personalizes and executes accurately.

\section{Personalized Web Agent}
\label{sec:webagent}
\subsection{Architecture}
\label{subsec:model}
To investigate how well existing web agents can perform personalized web navigation through \proposed, we conduct experiments using two web agent architectures, AgentOccam~\cite{Yang2024AgentOccamAS} and Browser-Use\footnote{\url{https://browser-use.com/}}.
(For more details about these architectures, see Appendix~\ref{appendix:webagent}.)
However, these architectures are designed for generic web navigation and do not account for user history by design, making them unsuitable for evaluating personalized web navigation tasks.
Therefore, we transform these architectures into the end-to-end personalized web navigation pipelines by augmenting them with a personalization module that retrieves and reasons over user history.
Specifically, our transformed architecture consist of three components: \textbf{planner}, \textbf{retriever}, and \textbf{generator}.

\begin{itemize}
[leftmargin=*,topsep=2pt,itemsep=2pt,parsep=0pt]
    \item \textbf{Planner:}
    The planner interprets the current observation and determines whether user history is needed. 
    If so, it generates a history query to retrieve relevant history; otherwise, it directly generates the execution plan for the next step.
    
    \item
    \textbf{Retriever:}
    The retriever receives history queries from the planner and searches for relevant entries in the user history.
    More details in Section~\ref{appendix:retriever}.
    
    \item
    \textbf{Generator:}
    The generator reasons over retrieved histories to resolve query ambiguities. 
    It receives the history query from the planner along with the retrieved histories from the retriever and analyzes them to identify preferences and extract specific details.
    It then produces a personalized rationale that guides subsequent action planning.
\end{itemize}

\noindent
Throughout the process, the agent generates a trajectory that captures both actions and reasoning at each step~\cite{Erdogan2025PlanandActIP}. 
The resulting reasoning trajectory documents the accessed history entries, query disambiguation process, and final option selection.

\subsection{History Access Scheme}
\label{subsec:accessscheme}

To evaluate distinct aspects of personalization capabilities, we propose two history access schemes for web agents that differ in when and how user history is accessed: \textbf{on-demand} and \textbf{pre-execution}.
In the \textbf{on-demand} scheme, the agent accesses user history dynamically during task execution, only when the planner determines it is needed at each step.
This scheme evaluates whether the agent can recognize, in real time, when and what type of personalization is required, thereby testing its situational awareness.
In the \textbf{pre-execution} scheme, the agent retrieves all relevant histories before execution begins by generating multiple queries to resolve ambiguities.
This scheme examines the agent's ability to engage in long-horizon planning to anticipate what information will be needed during execution.
Refer to Appendix~\ref{appendix:historyscheme} for detailed analysis.
Full trajectory examples for each scheme are provided in Table~\ref{tab:pre_example} and Table~\ref{tab:on_example}.

\begin{table*}[t]
\caption{Performance comparison across two agent architectures (AgentOccam, Browser Use) with five backbone LLMs under three history access schemes (No-history, On-demand, Pre-execution). Metrics include $\pweb$, $\ppref$, $\pavg$(average of $\pweb$ and $\ppref$), Intent satisfaction (Intent), and Success Rate (SR). For No-history scheme, we report only $\pavg$ for brevity.}
\label{tab:main_compact_reordered}
\centering
\small
\renewcommand{\arraystretch}{1.1}

\newcolumntype{P}{>{\centering\arraybackslash}p{0.04\textwidth}}

\resizebox{0.99\textwidth}{!}{%
\begin{tabular}{lPPP PPPPP PPPPP}
\toprule
\multirow{2.5}{*}{\textbf{Model}}
& \multicolumn{3}{c}{\textbf{No-history}}
& \multicolumn{5}{c}{\textbf{On-demand}}
& \multicolumn{5}{c}{\textbf{Pre-execution}} \\
\cmidrule(lr){2-4}\cmidrule(lr){5-9}\cmidrule(lr){10-14}

& $\pavg$ & Intent & SR
& $\pweb$ & $\ppref$ & $\pavg$ & Intent & SR
& $\pweb$ & $\ppref$ & $\pavg$ & Intent & SR \\
\midrule

\rowcolor{gray!15}[0pt][0pt]
\multicolumn{14}{@{}l@{}}{\textbf{AgentOccam}} \\

o3
& 0.168 & 0.503 & 0.00
& \textbf{0.747} & 0.551 & 0.649 & \textbf{0.423} & \textbf{0.07}
& 0.683 & 0.555 & 0.619 & 0.397 & \textbf{0.07} \\

GPT-4.1
& 0.126 & 0.503 & 0.00
& 0.740 & \textbf{0.577} & \textbf{0.658} & 0.417 & 0.06
& \textbf{0.705} & 0.574 & 0.639 & \textbf{0.409} & \textbf{0.07} \\

Gemini-2.5-Flash
& 0.118 & 0.500 & 0.00
& 0.603 & 0.368 & 0.486 & 0.403 & 0.02
& 0.517 & 0.208 & 0.362 & 0.377 & 0.00 \\

Qwen3-80B-Inst.
& 0.122 & 0.463 & 0.00
& 0.620 & 0.480 & 0.550 & 0.453 & 0.07
& 0.703 & \textbf{0.646} & \textbf{0.675} & 0.347 & 0.05 \\

Llama-3.3-70B
& 0.130 & 0.242 & 0.00
& 0.683 & 0.491 & 0.587 & 0.140 & 0.01
& 0.693 & 0.552 & 0.622 & 0.217 & 0.00 \\

\midrule

\rowcolor{gray!15}[0pt][0pt]
\multicolumn{14}{@{}l@{}}{\textbf{Browser-Use}} \\

o3
& 0.034 & 0.277 & 0.00
& 0.743 & 0.568 & 0.655 & \textbf{0.530} & \textbf{0.13}
& 0.733 & 0.608 & 0.671 & \textbf{0.450} & 0.10 \\

GPT-4.1
& 0.077 & 0.250 & 0.00
& \textbf{0.803} & \textbf{0.731} & \textbf{0.767} & 0.467 & \textbf{0.13}
& \textbf{0.767} & \textbf{0.686} & \textbf{0.727} & 0.530 & \textbf{0.13} \\

Gemini-2.5-Flash
& 0.095 & 0.270 & 0.00
& 0.707 & 0.487 & 0.597 & 0.283 & 0.02
& 0.757 & 0.561 & 0.659 & 0.350 & 0.03 \\

Qwen3-80B-Inst.
& 0.098 & 0.230 & 0.00
& 0.797 & 0.551 & 0.674 & 0.297 & 0.03
& 0.760 & 0.680 & 0.720 & 0.407 & 0.03 \\

Llama-3.3-70B
& 0.078 & 0.090 & 0.00
& 0.690 & 0.534 & 0.612 & 0.190 & 0.01
& 0.713 & 0.647 & 0.680 & 0.197 & 0.02 \\

\bottomrule
\end{tabular}%
}
\end{table*}

\section{Experiments}
\label{sec:experiments}
\subsection{Meta Evaluation}
\label{subsec:metaeval}
\begin{table}[!t]
\caption{{Meta evaluation results comparing reasoning-aware (ours), action-wise, and outcome-based evaluation.}}
\label{tab:meta_eval}
\centering
\renewcommand{\arraystretch}{1.12}

\resizebox{\columnwidth}{!}{
\begin{tabular}{llccc}
\toprule
\textbf{Method} & \textbf{Eval metric} & \textbf{Pearson} & \textbf{Spearman} & \textbf{Accuracy} \\
\midrule

\multirow{3}{*}{\parbox[t]{0.35\linewidth}{\textbf{Reasoning-aware (ours)}}}
& Website     & 0.8525 & 0.8611 & 0.8400 \\
& Preference  & 0.7200 & 0.7248 & 0.8800 \\
& Intent      & 0.7386 & 0.7436 & 0.7400 \\
\midrule

\multirow{3}{*}{Action-wise}
& Website     & 0.7654 & 0.8163 & 0.8400 \\
& Preference  & 0.3993 & 0.4026 & 0.5600 \\
& Intent      & 0.6409 & 0.6330 & 0.6800 \\
\midrule

\multirow{3}{*}{\parbox[t]{0.35\linewidth}{Outcome-based}}
& Website     & 0.8038 & 0.8209 & 0.8300 \\
& Preference  & 0.2253 & 0.2241 & 0.4600 \\
& Intent      & 0.5909 & 0.5933 & 0.7000 \\
\bottomrule
\end{tabular}
}
\end{table}
To validate our evaluation framework, we conduct a meta-evaluation on 50 newly generated queries by comparing its correlation with human judgments.
\vspace{-0.5em}
\paragraph{Baselines and settings.}
Using GPT-4.1 as the backbone model, we instruct the agent to perform tasks under both pre-execution and on-demand schemes.
GPT-5-mini serves as the meta evaluator, consistent with our main experiment.
We compare our method against two other commonly adopted methods: \textit{action-wise}, which evaluates the action sequence across all steps, and \textit{outcome-based}, which considers only the final action and its reasoning.
Since these two methods lack access to intermediate history retrieval steps, they evaluate only whether the ground truth preferences are reflected in the agent's actions when measuring PS.
This corresponds to the \textit{utilization accuracy} of our full rubric.
\vspace{-0.5em}
\paragraph{Evaluation process.}
Human annotators evaluate each pair of trajectory by determining which scheme performs better (pre-execution, on-demand, or tie) across preference, website, and intent, with ground truth values provided for reference.
To compare our evaluation framework against human judgment, we convert our scores into pairwise outcomes indicating which scheme received a higher score for each task.
We then calculate Pearson and Spearman correlation coefficients, as well as accuracy, between each evaluation method and human annotations across all three metrics to quantify alignment with human judgment.
For the reliability of the human-annotated reference, we additionally report inter-annotator agreement in Appendix~\ref{appendix:judge_validation}.

\paragraph{Evaluation results.}
In Table~\ref{tab:meta_eval}, our reasoning-aware method achieves the highest correlation with human judgments across all metrics.
Notably, ours and the action-wise method both evaluate full trajectories, but ours incorporates reasoning traces. 
This difference yields a significant performance gap, confirming that reasoning is essential for accurately evaluating personalization.
Beyond this alignment with human judgment, the LLM judge's internal consistency and agreement with alternative judges are further examined in Appendix~\ref{appendix:judge_validation}.

\subsection{Experimental Settings}
We assess how personalization performance varies across agent architectures, backbone LLMs, and history access schemes.
Specifically, we augment two base architectures (AgentOccam, Browser-Use) with our personalization module (Section~\ref{subsec:model}) under three history access schemes (Section~\ref{subsec:accessscheme}).
We evaluate performance using five metrics: $\pweb$, $\ppref$, $\pavg$, intent satisfaction, and success rate.

\subsection{Main Results}
\label{subsec:mainresult}
We evaluate personalized web agents with five backbone LLMs: o3~\cite{openai2025o3systemcard}, GPT-4.1~\cite{openai2024gpt41systemcard}, Gemini-2.5-Flash~\cite{google2024gemini25flash} for proprietary models, Qwen3-80B-Instruct~\cite{qwen3_2025}, and Llama-3.3-70B-Instruct~\cite{meta2024llama3} for open-source models.
All experiments use level 2 queries, which are the most ambiguous, under three history access schemes.
Results are presented in Table~\ref{tab:main_compact_reordered}.
We further examine the reproducibility of our results under repeated executions and temporal shifts in Section~\ref{subsec:openwebreproducibility}.

\paragraph{Performance differs across agent architectures and backbone models.}
When history access is enabled (on-demand or pre-execution), Browser-Use consistently achieves higher personalization scores ($\pweb$, $\ppref$) than AgentOccam across all backbone models;
this indicates that Browser-Use more effectively applies inferred user preferences during execution. 
We attribute this gap to differences in observation construction.
AgentOccam prunes the accessibility tree to retain only pivotal nodes and their related elements, producing a condensed representation that may discard relevant contexts.
Browser-Use, in contrast, constructs a full enhanced DOM tree that preserves richer information (Appendix~\ref{appendix:webagent}).

Across both agent architectures, proprietary models (o3 and GPT-4.1) substantially outperform open-source models, achieving success rates of 6--13\% under history access, while most open-source models remain below 5\%. 
Among open-source backbones, Qwen3-80B-Instruct is the only model that approaches proprietary-level performance in specific settings. 
These results suggest that both agent architecture and backbone model capacity jointly constrain personalization performance.
\vspace{-0.5em}
\paragraph{Web agents cannot handle ambiguous queries without user history.}
Under no-history scheme, all models across both agents achieve 0\% success.
This confirms that web agents cannot resolve ambiguous queries without access to user history. 
Notably, the two agents exhibit distinct failure patterns. 
AgentOccam achieves approximately 2$\times$ higher intent satisfaction than Browser-Use by substituting missing values with random values, prioritizing task completion over accuracy.
In contrast, Browser-Use requests user clarification and terminates execution, prioritizing accuracy over completion but resulting in task abandonment. 
In both cases, ambiguity remains unresolved.
This inability to handle ambiguity cannot be captured through existing benchmarks that provide explicit queries, which highlights the necessity of our \textit{clarify-to-personalize} queries and user history.
\vspace{-0.5em}
\paragraph{Task completion alone cannot measure personalization capability.}
For personalized web agents, task success depends on both personalization and navigation, and weakness in either leads to failure.
For instance, Browser-Use with Llama-3.3-70B under pre-execution achieves $\pavg=0.680$ but only $\text{Intent}=0.197$, which results in $\text{SR}=0.02$. 
Conversely, AgentOccam with Gemini-2.5-Flash under on-demand achieves $\text{Intent}=0.403$ but only $\pavg=0.486$, which also results in $\text{SR}=0.02$.
Both cases record the same low success rate, yet they fail for opposite reasons.
Llama succeeds at personalization but struggles with navigation, while Gemini navigates reasonably but fails at personalization.
That is, task completion alone cannot distinguish whether failures arise from incorrect personalization or ineffective navigation.
Our reasoning-aware evaluation framework resolves this by evaluating personalization and navigation separately.
\vspace{-0.5em}
\paragraph{History access schemes reveal backbone strengths.}
Some backbone models consistently perform better under a specific scheme regardless of agent architecture.
For instance, GPT-4.1 consistently achieves higher $\pavg$ under on-demand than pre-execution across both agents, while Qwen3-80B-Instruct and Llama-3.3-70B show the opposite pattern, performing relatively better under pre-execution.
This indicates that the two schemes capture different capabilities of these models as mentioned in Section~\ref{subsec:accessscheme}.
Based on this result, GPT-4.1 appears better suited for situational awareness, while Qwen3-80B-Instruct and Llama-3.3-70B are for long-horizon planning.

\begin{figure}[t]
    \centering
    \includegraphics[width=\columnwidth]{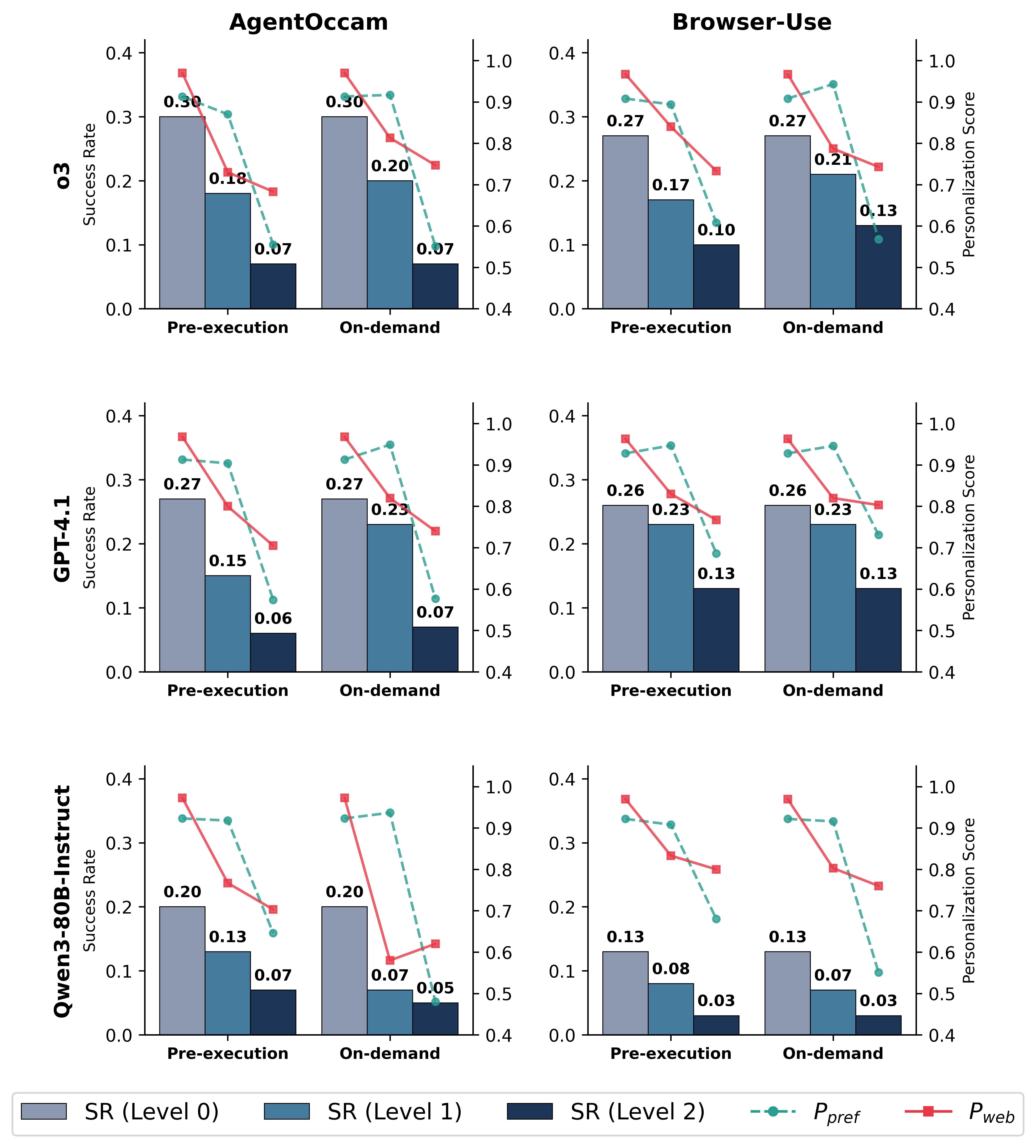}
    \caption{Performance across query ambiguity levels for each agent architecture and history access scheme. Bars show Success Rate (left axis) for level 0, 1, 2 (gray, medium, dark blue). Lines show preference (red) and website (green) scores (right axis). o3, GPT-4.1, and Qwen3-80B-Instruct are used as backbone models.}
    \label{fig:query_amb}
\end{figure}

\subsection{Analysis Across Query Ambiguity}
\label{subsec:analysis}
This section analyzes how performance changes across query ambiguity levels for each combination of agent architecture, backbone models, and two history access schemes.
Based on the main results presented in ~\ref{subsec:mainresult}, we select the top three performing backbone models--o3, GPT-4.1, and Qwen3-80B-Instruct--for the analysis.
Figure~\ref{fig:query_amb} reports success rate and personalization scores across different ambiguity levels (level 0--2).

\paragraph{{Performance degrades as ambiguity level increases.}}
Well-personalized web agents should maintain performance even on ambiguous queries by leveraging user history.
However, despite providing access to user history through our personalization module, success rate consistently decreases as query ambiguity increases across all configurations.
Clear queries achieve an average success rate of 23.8\%, while level 1 queries drop to 16.3\%, and level 2 to 7.8\%.
Since both the personalized architecture and user history are available, this performance drop suggests that agents struggle to effectively identify, reason over, and utilize relevant historical information to disambiguate queries as ambiguity increases.

\paragraph{Clear queries still lead to personalization failures.}
To identify factors limiting personalization capability, we analyze performance on clear queries, where all required information is explicitly provided and no retrieval is needed.
Yet, preference scores average only 0.92 and website scores 0.97 on clear queries, neither reaching a perfect 1.0.
This indicates that even when complete information is given, agents often fail at utilization.
We trace these failures to two primary causes.
First, agents struggle to handle multiple constraints, applying only a subset.
For example, given ``my usual external SSD capacity, port, and budget,'' agents correctly applied capacity (1TB-1.9TB) and port (USB interface) filters, but failed to apply the budget constraint (under \$120).
Second, agents fail to determine when and where to apply given information, specifically identifying the appropriate element or execution step remains challenging.
These findings indicate that improving personalization requires advancing not only retrieval but also utilization capabilities.

\subsection{Analysis On User Context Implicitness}
\label{subsec:analysis_usercontext}
\begin{table}[t]
\caption{Performance with explicit user profiles using Browser-Use. Values in parentheses indicate changes relative to implicit user history results in Table~\ref{tab:main_compact_reordered}.}
\label{tab:usercontext}
\centering
\small
\setlength{\tabcolsep}{8pt}
\resizebox{0.99\columnwidth}{!}{%
\begin{tabular}{l@{\hskip 4pt}lcc}
\toprule
\multirow{2.5}{*}{\shortstack{\textbf{History}\\\textbf{Scheme}}} &
\multirow{2.5}{*}{\textbf{Metric}} &
\multicolumn{2}{c}{\textbf{Backbone}} \\
\cmidrule(lr){3-4}
& & \textbf{o3} & \textbf{GPT-4.1} \\
\midrule
\multirow{4}{*}{Pre-execution} & $\pweb$ & 0.830 {\color{green!50!black}(+0.097)} & 0.887 {\color{green!50!black}(+0.120)} \\
& $\ppref$ & 0.764 {\color{green!50!black}(+0.156)} & 0.892 {\color{green!50!black}(+0.206)} \\
& Intent & 0.400 {\color{green!50!black}(+0.050)} & 0.640 {\color{green!50!black}(+0.110)} \\
& Success & 0.17 {\color{green!50!black}(+0.07)} & 0.27 {\color{green!50!black}(+0.14)} \\
\midrule
\multirow{4}{*}{On-demand} & $\pweb$ & 0.890 {\color{green!50!black}(+0.147)} & 0.933 {\color{green!50!black}(+0.130)} \\
& $\ppref$ & 0.682 {\color{green!50!black}(+0.114)} & 0.887 {\color{green!50!black}(+0.156)} \\
& Intent & 0.617 {\color{green!50!black}(+0.087)} & 0.597 {\color{green!50!black}(+0.130)} \\
& Success & 0.15 {\color{green!50!black}(+0.02)} & 0.25 {\color{green!50!black}(+0.12)} \\
\bottomrule
\end{tabular}%
}
\end{table}
We construct our user history to reveal preferences implicitly through behavioral patterns rather than explicit statements.
This is critical since explicit statements reduce personalization to instruction-following, where agents simply apply given information, not inferring user intent.
This undermines the genuine evaluation of personalization.
To validate our implicit encoding and its necessity for personalization evaluation, we compare performance when given \textbf{explicit user profile} (Section~\ref{subsec:userhistory}), which explicitly includes preference statements, and \textbf{implicit user history}, which implicitly reveals preferences via behavioral patterns. 
In Table~\ref{tab:usercontext}, all configurations achieve substantially higher performance across all metrics with explicit profiles.
Both contexts contain the same user information, yet there exists significant performance gap.
This confirms that preferences in our user history are not explicitly visible but must be inferred from behavioral patterns, validating our implicit encoding design.
Furthermore, it highlights the necessity of implicit user context for genuine personalization and evaluation.

\subsection{Reproducibility under Open-web Dynamics}
\label{subsec:openwebreproducibility}
\begin{table}[t]
\centering
\caption{Repeated experiments over three runs using Browser-Use.}
\label{tab:repeated_runs}
\small
\begin{tabular*}{\columnwidth}{@{\extracolsep{\fill}}llcc}
\toprule
\textbf{Model} & \textbf{Metric} & \textbf{On-demand} & \textbf{Pre-execution} \\
\midrule
\multirow{4}{*}{GPT-4.1}
    & $\pweb$  & $0.834 \pm 0.015$ & $0.807 \pm 0.013$ \\
    & $\ppref$ & $0.790 \pm 0.005$ & $0.790 \pm 0.006$ \\
    & Intent  & $0.643 \pm 0.025$ & $0.559 \pm 0.015$ \\
    & Success & $0.022 \pm 0.000$ & $0.018 \pm 0.003$ \\
\midrule
\multirow{4}{*}{Qwen-80B}
    & $\pweb$  & $0.779 \pm 0.005$ & $0.807 \pm 0.007$ \\
    & $\ppref$ & $0.560 \pm 0.003$ & $0.774 \pm 0.008$ \\
    & Intent  & $0.301 \pm 0.012$ & $0.448 \pm 0.025$ \\
    & Success & $0.008 \pm 0.001$ & $0.012 \pm 0.001$ \\
\bottomrule
\end{tabular*}
\end{table}
\begin{table}[t]
\centering
\caption{Temporal variance between December 2025 and March 2026 runs using Browser-Use.}
\label{tab:temporal_variance}
\small
\setlength{\tabcolsep}{3.5pt}
\begin{tabular*}{\columnwidth}{@{\extracolsep{\fill}}llrrrr}
\toprule
\textbf{Model} & \textbf{Scheme} & \textbf{$\pweb$} & \textbf{$\ppref$} & \textbf{Intent} & \textbf{Success} \\
\midrule
\multirow{2}{*}{GPT-4.1} & On     
    & $+0.031$ & $+0.059$ & $+0.176$ & $-0.108$ \\
        & Pre
    & $+0.040$ & $+0.104$ & $+0.029$ & $-0.112$ \\
\midrule
\multirow{2}{*}{Qwen-80B} & On     
    & $-0.018$ & $+0.009$ & $+0.004$ & $-0.022$ \\
         & Pre 
    & $+0.047$ & $+0.094$ & $+0.041$ & $-0.018$ \\
\bottomrule
\end{tabular*}
\end{table}

Since agents on the open web encounter dynamic environments (e.g., different search results, item availability) across executions~\cite{Pan2024WebCanvasBW, Rawles2024AndroidWorldAD, He2024WebVoyagerBA}, reliable evaluation must distinguish agent behavior from such variability.
We therefore examine whether \proposed\ yields stable scores under open-web dynamics in two dimensions: execution variability and temporal variability.
Execution variability measures score variation across repeated runs under similar time conditions, while temporal variability measures variation across runs collected several months apart.
We conduct both analyses on a representative subset, using Browser-Use with the best performing proprietary and open-source backbones in Table~\ref{tab:main_compact_reordered}, GPT-4.1 and Qwen-80B-Instruct, respectively.
\vspace{-0.5em}
\paragraph{Execution variability}
Table~\ref{tab:repeated_runs} shows that three independent runs produce low variance, with a maximum standard deviation of 0.025.
This indicates stable scoring under repeated open-web execution.
\vspace{-0.5em}
\paragraph{Temporal variability}
Table~\ref{tab:temporal_variance} reports the differences between the scores in Table~\ref{tab:main_compact_reordered} and~\ref{tab:repeated_runs}, obtained three months apart.
Most metrics exhibit small absolute differences between the two periods.
The relative model ranking also remains unchanged over time, as GPT-4.1 continues to outperform Qwen-80B-Instruct across schemes and metrics.

Overall, these findings indicate that the results in Table~\ref{tab:main_compact_reordered} remain consistent despite open-web dynamics, supporting the faithfulness of the evaluation.
Our evaluation pipeline is based on reasoning traces rather than final outcomes, and allows partial credit when external factors block correct actions that would otherwise succeed.
This prevents open-web variability from being conflated with agent capability.

\section{Error Analysis}
\label{sec:erroranalysis}
To investigate the sources of personalization failures, we conduct an analysis based on the agent's trajectories from level 2 query experiments. 
We examine them with imperfect PS and categorize observed errors into four distinct types: {Redundant history access}, {Personalization hallucination}, {History retrieval failure}, and {History utilization failure}. 
Beyond the qualitative analysis, Appendix~\ref{appendix:error} provides complementary error statistics across backbone models.

\paragraph{Redundant history access.}
Redundant history access occurs when the agent generates underspecified personalization queries, leading to unsuccessful retrieval and repeated attempts.
In Table~\ref{tab:error1}, the agent queries ``\textit{user's usual shore excursion preferences},'' which is too broad to identify relevant history.
When the query is refined by including domain-aware attributes ``\textit{activity type, activity level, price range}'', the agent retrieves relevant history.
Without such specificity, the agent repeatedly accesses user history assuming no relevant history exists, causing redundant retrieval attempts.

\paragraph{Personalization hallucination.}
Personalization hallucination refers to errors in which the model fabricates user information without accessing relevant histories. 
In Table~\ref{tab:error3}, the agent incorrectly applies a non-existent preference \textit{4 stars and up} as if derived from the user history. 
Such errors frequently occur with rating or pricing thresholds, where the agent defaults to generic assumptions (e.g., preferring higher ratings or lower prices) instead of referencing user history. 
Although such behavior may appear operationally sound, it contradicts the goal of personalization.
The goal is not to generate plausible information, but to reason faithfully from user-grounded evidence. 

\paragraph{History retrieval failure.}
History retrieval failure refers to cases in which the agent retrieves irrelevant or incomplete histories, leading to the omission of essential information.
In Table~\ref{tab:error2}, the agent fails to identify ``\textit{Baltimore Orioles}'' as the user's favorite MLB team. 
Although it is mentioned in multiple histories, none of them explicitly state ``favorite team'' causing the retriever to miss the connection.
To understand whether retriever design contributes to such failures, we explore alternative indexing designs in Appendix~\ref{appendix:retrieveridx}.

\paragraph{History utilization failure.}
History utilization failure represents situations where appropriate history is successfully retrieved but not properly utilized in the agent’s action. 
In Table~\ref{tab:error4}, the agent retrieves the user’s location \textit{Los Angeles} but fails to incorporate it into searching, resulting in incorrect results. 
This failure stems from the agent’s inability to determine when the retrieved histories should be applied within the overall process. 
The location needs to be set prior to initiating the search to yield accurate results.

\section{Conclusion}
\label{sec:conclusion}
This paper establishes the first benchmark for evaluating personalized web agents in real open web environments. 
Our benchmark introduces a \textit{clarify-to-personalize} query design and rigorously constructs user histories, along with a reasoning-aware evaluation framework that assesses personalization through rubric-based scoring. 
We conduct extensive experiments across web agent architectures, multiple backbone LLMs, two history access schemes, query ambiguity, and user context formats, followed by error analysis to identify key challenges in web agent personalization.
By enabling the assessment of how agents infer and apply user preferences, we believe this benchmark provides a strong foundation for advancing personalization in web agents.




\section*{Acknowledgements}

This work was supported by the IITP grants funded by the Korea government (MSIT) (RS-2024-00457882, AI Research Hub Project; RS-2026-25520654).

\section*{Impact Statement}

This paper presents work whose goal is to advance the field of personalized web agents. 
Our \proposed\ uses synthetically generated user profiles and browsing histories, ensuring no real user data is collected or used. 
However, we acknowledge that deployed personalized agents would require access to actual user behavior patterns, raising privacy considerations regarding consent and data protection. 
We encourage future work to carefully address these concerns. 
Additionally, as our benchmark operates on the real open web, agents may encounter potentially harmful content during execution. 
We hope our work encourages research into personalization approaches that incorporate both privacy safeguards and safety mechanisms for web navigation.


\bibliography{reference}
\bibliographystyle{icml2026}

\newpage
\appendix
\label{sec:appendix}
\section{Dataset}
\label{appendix:dataset}

\subsection{Dataset Details}
\label{appendix:datasetdetails}

\begin{figure}[h]
    \centering
    \includegraphics[width=\columnwidth]{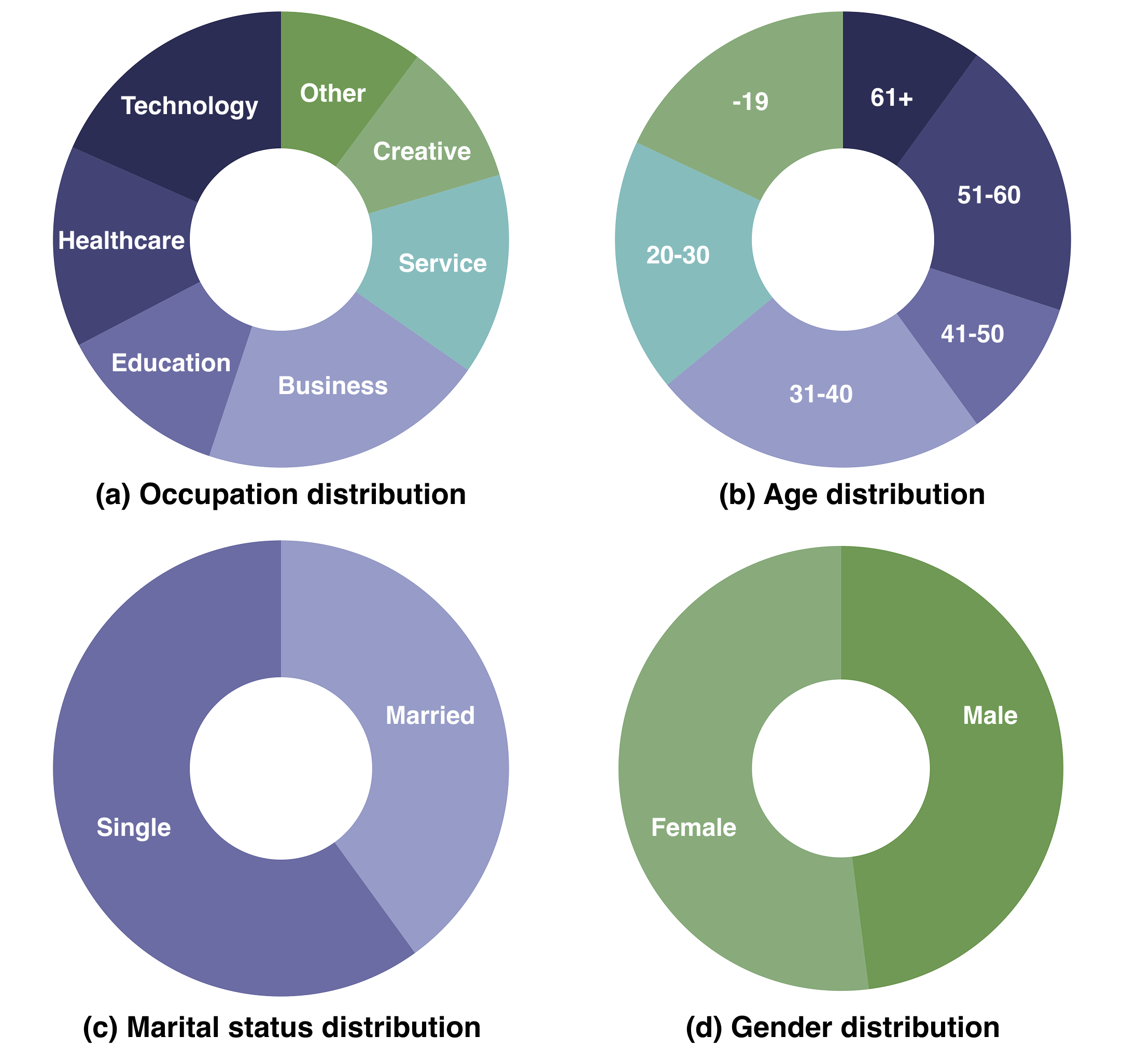}
    \caption{Detailed statistics of user profile provided from \proposed\ benchmark.}
    \label{fig:statisticsuserprofile}
\end{figure}

We construct a user profile of 50 individuals whose occupation, age, gender, and marital status are evenly distributed to ensure demographic balance, as depicted in Figure 4. 
Each attribute combination is designed to be plausible and internally consistent, avoiding unrealistic pairings such as a 16 years old professor or a 21 years old retiree. 
To ensure this, both human validation and prompt-based filtering are employed during generation. 
The occupation categories are defined across six major sectors—technology, healthcare, education, business, service, and creative industries. 
To define realistic web domains, we collected 144 websites used in prior benchmarks (Mind2Web, WebArena, WebVoyager) and filtered out sites with CAPTCHAs, regional restrictions, or outdated content, resulting in 105 accessible websites.
We categorized these into 21 representative domains: Groceries, Apparel and accessories, Electronics, Home and kitchen, Books and stationary, Sports and outdoor, Transportation, Hotels, Car rental, Tours, Restaurants reservations, Weather and maps, Health and medical, Movies, Music, Concerts and exhibitions, Games, Online learning, Search, and Communities and forums.
The inclusion of diverse occupations and domains results in heterogeneous user profiles, each exhibiting distinct interests and behavioral routines, thereby enabling the generation of diverse and realistic user interaction histories.
We construct approximately 2,000 history entries spanning a one-year period for each of the 50 users.

\subsection{Dataset Validation}
\label{appendix:datasetvalidation}
\paragraph{User Profile}
We employed GPT-5 during data construction~\cite{Zhao2025DoLR, kwak2025toolhaystackstresstestingtoolaugmentedlanguage} to verify whether all the generated event seeds and decomposed actions adhere to the generation instructions (Table 8-10) and to detect errors (e.g., formatting error, temporal inconsistency, preference misalignment, etc).
All identified issues were corrected before generating the user histories.

Human annotators additionally reviewed each user profile against five criteria: (a) plausibility and internal consistency of demographic field combinations, (b) compatibility between preferences/behavioral patterns and demographic characteristics, (c) whether selected websites actually support the described functionalities, (d) whether behavioral patterns exhibit incoherent mixtures of unrelated routines, and (e) absence of conflicts between preferences and behavioral patterns.
Profiles failing any criterion were revised and re-examined until all were satisfied.

\begin{table}[h]
\centering
\caption{Diversity of user histories across global, cross-user, and per-user scales.}
\label{tab:diversity_statistics}
\small
\begin{tabular}{l l c}
\toprule
\textbf{Scale} & \textbf{Metric} & \textbf{Value} \\
\midrule
\multirow{2}{*}{Global}
    & Topic entropy & 4.503 \\
    & Shannon evenness     & 0.978 \\
\midrule
\multirow{2}{*}{Cross-user}
    & Cosine distance & 0.626 \\
    & Self-BLEU       & 0.235 \\
\midrule
Per-user
    & Ent-4 & 9.932 \\
\bottomrule
\end{tabular}
\end{table}
\paragraph{Diversity of User History}
\proposed's user histories must be sufficiently diverse to reveal distinct preferences across users and rich behavioral patterns within each user.
We therefore validate user history diversity at three scales: global (overall topical coverage), cross-user (inter-user distinctiveness), and per-user (intra-user variation).
In addition to standard lexical diversity metrics (Ent-4, Self-BLEU, Shannon evenness) used in \citet{Joko2024DoingPL, Zhao2025PersonaLensAB}, topic clustering over sentence embeddings are adopted to capture semantic diversity beyond lexical variation.
Results are reported in Table~\ref{tab:diversity_statistics}.

At the global scale, topic entropy and Shannon evenness measure how broadly and uniformly history entries from all users span topic clusters obtained from sentence embeddings.
The topic entropy of $4.503$ (out of a maximum of $4.605$) and an evenness of $0.978$ together indicate broad coverage without any dominant subset.
At the cross-user scale, the cosine distance between users' topic distributions and the Self-BLEU between users' entries quantify topical and textual distinctiveness. 
The result attains $0.626$ and $0.235$ respectively, proposing the contextual distinctiveness with low textual overlap.
At the per-user scale, Ent-4 measures the entropy of 4-gram distributions within each individual history. 
The Ent-4 of $9.932$ exceeds the prior leading score of $8.72$~\cite{Zhao2025PersonaLensAB}, which demonstrates the high lexical diversity within each user's history.

\begin{table}[h]
\centering
\caption{Pairwise human evaluation of user history realism. Win Rate denotes the proportion of judgments where \proposed\ is preferred over histories generated without our pipeline. Gwet's AC1 measures IAA among annotators.}
\label{tab:dataset_realism}
\small
\begin{tabular}{l c c}
\toprule
\textbf{Level} & \textbf{Win Rate} & \textbf{Gwet's AC1} \\
\midrule
History sequence & 88.3\% & 0.72 \\
History entry    & 98.3\% & 0.97 \\
\bottomrule
\end{tabular}
\end{table}
\paragraph{Realism of User History}
To assess whether the generated histories reflect plausible human behavioral patterns, we conducted a pairwise human evaluation against histories generated without our multi-stage pipeline.
Six human judges were provided with the pair of each profile and asked to select which one more faithfully represents the web activity of the given user.
The evaluation covers two dimensions.
The first, \textit{history sequence}, examines whether the action sequence reflects realistic behavioral flow.
The second, \textit{history entry}, captures how clearly individual entries manifest user-specific preferences.
Our dataset achieved win rates of 88.3\% and 98.3\% at the sequence level and entry level, respectively.
Furthermore, human inter-annotation agreement (IAA) measured by Gwet's AC1 reached 0.72 and 0.97 respectively, indicating strong agreement.

\section{LLM Judge Validation}
\label{appendix:judge_validation}

\subsection{Human Inter-annotator Agreement}
\begin{table}[h]
\centering
\caption{Human inter-annotator agreement in meta-evaluation.}
\label{tab:human_iaa}
\small
\begin{tabular}{lccc}
\toprule
\textbf{Metric} & \textbf{$\pweb$} & \textbf{$\ppref$} & \textbf{Intent} \\
\midrule
Average agreement & 100.0 & 92.0 & 81.3 \\
Gwet's AC1 & 1.000 & 0.902 & 0.743 \\
\bottomrule
\end{tabular}
\end{table}
In our meta-evaluation (Section~\ref{subsec:metaeval}), human judgments serve as the reference for evaluating how well each evaluation method aligns with humans.
To verify that the human reference is reliable, three annotators independently label the same set of trajectory pairs and measure IAA on each metric.
We report both average agreement, which measures the proportion of agreement between annotators, and Gwet's AC1, which adjusts this agreement for chance.
As shown in Table~\ref{tab:human_iaa}, annotators reach perfect agreement on $\pweb$, near perfect agreement on $\ppref$ (AC1 $=0.902$), and substantial agreement on intent (AC1 $=0.743$).
This supports the faithfulness of human judgments as a reference for assessing our evaluation framework.

\subsection{Internal Consistency}
\begin{table}[h]
\centering
\caption{Average and standard deviation across three repeated evaluations of GPT-5-mini on Browser-Use with GPT-4.1 trajectories.}
\label{tab:internal_consistency}
\small
\begin{tabular}{lccc}
\toprule
\textbf{Scheme} & \textbf{$\pweb$} & \textbf{$\ppref$} & \textbf{Intent} \\
\midrule
On     & $0.828 \pm 0.006$ & $0.788 \pm 0.003$ & $0.619 \pm 0.002$ \\
Pre & $0.811 \pm 0.006$ & $0.771 \pm 0.012$ & $0.556 \pm 0.015$ \\
\bottomrule
\end{tabular}
\end{table}
We validate that our LLM judge (GPT-5-mini) produces consistent scores by evaluating the same trajectories (Browser-Use with GPT-4.1) three times and computing the variance of each metric across the three evaluations.
The success rate is excluded since it is computed rule-based.
Table~\ref{tab:internal_consistency} presents stable scores with standard deviations no greater than $0.015$ across all metrics and history access schemes, confirming the reliability of our LLM judge.

\subsection{Inter-judge Agreement}
\begin{table}[h]
\centering
\caption{Inter-judge agreement across three LLM judges (GPT-5-mini, Qwen-80B-Instruct, GPT-OSS-120B) on Browser-Use with GPT-4.1 trajectories.}
\label{tab:inter_judge}
\small
\begin{tabular}{llccc}
\toprule
\textbf{Agreement} & \textbf{Scheme} & \textbf{$\pweb$} & \textbf{$\ppref$} & \textbf{Intent} \\
\midrule
\multirow{2}{*}{Krippendorff's $\alpha$}
& Pre & 0.837 & 0.788 & 0.676 \\
& On     & 0.801 & 0.735 & 0.615 \\
\midrule
\multirow{2}{*}{ICC(A,1)}
& Pre & 0.839 & 0.789 & 0.679 \\
& On     & 0.802 & 0.737 & 0.608 \\
\bottomrule
\end{tabular}
\end{table}
Beyond internal consistency, we additionally examine the robustness to the choice of judge models.
Table~\ref{tab:inter_judge} reports the agreement across three judges: GPT-5-mini, Qwen-80B-Instruct, and GPT-OSS-120B.
Two open-source judges are intentionally included to mitigate potential bias from relying solely on proprietary models.
To reduce variance, each score is calculated on average of three runs on the same trajectories (Browser-Use with GPT-4.1).
Krippendorff's $\alpha$~\citep{Chaudhary2024TowardsUT, Haldar2025RatingRS} and ICC(A,1)~\citep{Li2024LLMsasJudgesAC} are reported, interpreted following Krippendorff's scale and the guideline of \citet{Koo2016AGO}, respectively.
$\pweb$ and $\ppref$ reach the reliable threshold ($\alpha \geq 0.800$) and the excellent ICC level ($\geq 0.75$) in most conditions, while intent remains within the tentatively acceptable range.
These results demonstrate that our evaluation framework is robust to the choice of judge model.

\section{Implementation Details}
\label{appendix:Implementation}

\subsection{Web Agent}
\label{appendix:webagent}
\paragraph{AgentOccam}
AgentOccam~\cite{Yang2024AgentOccamAS} improves web navigation performance without additional training by simply aligning the observation and action spaces with the LLM’s inherent capabilities.
It simplifies action space to reduce distractions and improving focus on meaningful operations
For the observation space, it removes redundant textual elements that 
describe page layout or functionality to produce a condensed representation 
of each page. 
It then refines observation history selectively by identifying pivotal nodes on the accessibility tree, which are web elements essential for task completion, and retaining these nodes along with their ancestors, siblings, and descendants. 
It then organizes the workflow through a planning tree to keep the agent focused. 
Each branch represents a new sub-goal and once a branch is initiated, past trajectories irrelevant to the current plan are pruned.
Following AgentOccam, we use a Playwright-based environment with Chromium as the browser engine. 
Unlike the original AgentOccam setup, we set the starting page to bing.com to avoid CAPTCHA issues.

\paragraph{Browser-Use}
Browser-Use extracts web page state through the Chrome DevTools Protocol (CDP). 
It retrieves three types of information in parallel: the DOM tree, layout information(bounding boxes and computed CSS styles), and the accessibility tree containing semantic role information for each element.
These three data sources are merged into an enhanced DOM tree where each node contains its structural properties, screen coordinates, visibility status, scrollability, and accessibility semantics. 
The framework then serializes this enhanced tree into a text representation for LLM input, assigning numeric indices to interactive elements. 
Browser-Use automatically switches to duckduckgo.com whenever it encounters CAPTCHAs.

While AgentOccam selectively retains pivotal nodes and their related elements 
to produce a condensed observation, Browser-Use constructs the full DOM tree and filters elements based on visibility and interactivity derived 
from the CDP snapshot data.
For fair comparison, we use only text-based observations without screenshot images.

\subsection{Retriever}
\label{appendix:retriever}
As a retriever, we use Stella V5 1.5B, the dense retriever that has shown high performance on MTEB~\cite{muennighoff-etal-2023-mteb}. 
Each history entry is indexed using the concatenation of its \textit{type} and \textit{object} attributes as the key, while the complete history item is stored as the value. 
We employ FAISS with inner product similarity (IndexFlatIP) for efficient nearest neighbor search. 
At retrieval time, we compute cosine similarity between the query embedding and all indexed keys, returning up to 20 entries that exceed a similarity threshold of 0.5. 
The embedding cache and FAISS index are serialized and stored on disk to enable fast loading across multiple task executions.

\section{History Access Schemes}
\label{appendix:historyscheme}
Our history access schemes evaluate fundamentally different dimensions of personalized web agent.

In the \textbf{on-demand} scheme, our analysis targets the agent’s capacity to detect \textit{when} personalization is required and \textit{what} kind of user information is needed based on the current state, not compensating for missing inputs with generic or randomly imputed values. 
The focus is thus not merely on end-to-end completion, but on whether the agent adheres to the \textit{clarify-to-personalize} goal, which is demonstrated by requesting user context via personalization module precisely when the current step demands it.
In practice, web agents cannot assume users are available to answer queries at every step of execution. 
In such environments, well-personalized web agents must be able to proactively access to the history when it is needed. 
On-demand evaluation directly probes real-time situational awareness and the necessity of timely personalization intervention under realistic constraints.

\textbf{Pre-execution} requires the agent to \textit{anticipate} what user information will be needed based on the given intent and domain, then \textit{integrate} retrieved histories to construct an enhanced query that resolves ambiguities. 
Anticipation demands long-horizon planning, while integration requires accurately synthesizing information across multiple domains. 
Unlike the on-demand scheme, history access is unavailable during navigation. 
Thus, any omission of constraints, inclusion of irrelevant histories, or reliance on stale information propagates throughout the entire trajectory.

\section{Evaluation Rubric}
\label{appendix:rubric}
We employ GPT-5-mini to evaluate agent performance.
The model receives the agent's full trajectory, including reasoning traces, and measures personalization and intent scores based on the rubrics.
Below, we detail the scoring criteria for each metric.

\subsection{Website Score}
The website score $\pweb$ evaluates whether the agent correctly identifies and navigates to the user's preferred website based on user history.
Total score: 10 points (5 for retrieval + 5 for utilization).

\begin{tcolorbox}[
  breakable,
  colback=gray!5,
  colframe=gray!50,
  title=Website Personalization Rubric,
  boxsep=3pt,
  left=2pt,right=2pt,top=2pt,bottom=2pt,
  before skip=2pt,
  after skip=2pt,
]
\small
\textbf{Task Context:} The agent needed to identify and navigate to the correct website based on the task query. \\

\textbf{STEP 1 - Retrieval Evaluation:}
\begin{enumerate}
    \item Did the agent generate appropriate queries to find website preference/history?
    \item Was the expected website found in ANY of the memory responses?
\end{enumerate}

\textbf{STEP 2 - Utilization Evaluation:}
\begin{enumerate}
    \item Did the agent successfully navigate to the expected website?
\end{enumerate}

\textbf{Scoring:}
\begin{itemize}
    \item \textbf{STEP 1 - Retrieval}: 5 pts if criteria met, 0 otherwise
    \item \textbf{STEP 2 - Utilization}: 5 pts if criteria met, 0 otherwise
\end{itemize}
\end{tcolorbox}

\subsection{Preference Score}
The preference score $\ppref$ evaluates whether the agent correctly retrieves and utilizes user preferences.
Total score: 10 points (2.5 per preference $\times$ 2 preferences $\times$ 2 steps).
For location-aware websites, an additional 5 points for location personalization. (Total score: 15 points)
The total score is normalized: $\pweb = \frac{\text{raw score}}{\text{max score}}$.

\begin{tcolorbox}[
  breakable,
  colback=gray!5,
  colframe=gray!50,
  title=Preference Personalization Rubric,
  boxsep=3pt,
  left=2pt,right=2pt,top=2pt,bottom=2pt,
  before skip=2pt,
  after skip=2pt,
]

\small
\textbf{Task Context:} The agent needed to retrieve and apply user preferences for the given query. \\

\textbf{Target Preferences:}
\begin{itemize}
    \item Preference 1 - Category: \{pref1\_key\}, Expected Value: \{pref1\_value\}
    \item Preference 2 - Category: \{pref2\_key\}, Expected Value: \{pref2\_value\}
\end{itemize}

\textbf{STEP 1 - Retrieval Evaluation:}

For EACH preference:
\begin{enumerate}
    \item Check if any memory query asked about this preference category
    \item Check if the expected value appears in any memory response
\end{enumerate}

\textbf{STEP 2 - Utilization Evaluation:}

For EACH preference:
\begin{enumerate}
    \item Check if the expected value was correctly applied in the reasoning
    \item OR check if the agent at least attempted to apply the expected value
\end{enumerate}

\textbf{Scoring:}

For EACH preference:
\begin{itemize}
    \item \textbf{STEP 1 - Retrieval}: 2.5 pts if criteria met, 0 otherwise
    \item \textbf{STEP 2 - Utilization}: 2.5 pts if criteria met, 0 otherwise
\end{itemize}
\end{tcolorbox}

\subsubsection{Location Evaluation (Conditional)}

For websites requiring location-aware navigation (Amazon, Instacart, Resy, OpenTable, etc), we additionally evaluate location personalization using similar retrieval and utilization rubrics, each worth 2.5 points.
The total score is normalized: $\ppref = \frac{\text{raw score}}{\text{max score}}$.

\begin{tcolorbox}[
  colback=gray!5,
  colframe=gray!50,
  title=Location Personalization Rubric,
  boxsep=3pt,
  left=2pt,right=2pt,top=2pt,bottom=2pt,
  before skip=2pt,
  after skip=2pt
]

\small
\textbf{Task Context:} The agent visited a location-required website and needed to apply user location.

\textbf{Expected Location:} \{city, state, zip\_code\}

\textbf{STEP 1 - Retrieval Evaluation:}
\begin{enumerate}
    \item Did the agent query for location/address information from memory?
    \item Was location data (city, state, zip code, address, neighborhood) retrieved?
\end{enumerate}

\textbf{STEP 2 - Utilization Evaluation:}
\begin{enumerate}
    \item Was the location applied during navigation? (delivery address, search location, filters, etc.)
\end{enumerate}

\textbf{Scoring:}
\begin{itemize}
    \item \textbf{STEP 1 - Retrieval}: 2.5 pts if criteria met, 0 otherwise
    \item \textbf{STEP 2 - Utilization}: 2.5 pts if criteria met, 0 otherwise
\end{itemize}
\end{tcolorbox}

\subsection{Intent Satisfaction}

Evaluates task completion independent of personalization accuracy, with partial credit for external failures.
Total score: 10 points.

\begin{tcolorbox}[
  breakable,
  colback=gray!5,
  colframe=gray!50,
  title=Intent Satisfaction Rubric,
  boxsep=3pt,
  left=2pt,right=2pt,top=2pt,bottom=2pt,
  before skip=2pt,
  after skip=2pt,
]

\small

\vspace{0.5em}
\textbf{10 POINTS - Complete Intent Satisfaction:}
\begin{itemize}
    \item The requested final goal was successfully achieved
    \item Examples: item added to cart, booking completed, information found
\end{itemize}

\textbf{5 POINTS - Critical Action Attempted but Blocked:}
\begin{itemize}
    \item Agent reached the correct website/service
    \item Agent identified the correct item/service
    \item Agent attempted the final decisive action (e.g., clicked ``Add to Cart'')
    \item BUT was blocked by external factors: out of stock, CAPTCHA, website error, service unavailable
\end{itemize}

\textbf{0 POINTS - Intent Not Satisfied:}
\begin{itemize}
    \item Failed to reach the decisive action
    \item Search/filtering failed
    \item Wrong item selected
    \item Process abandoned
    \item Constraint violated (e.g., attempted login when prohibited)
\end{itemize}
\end{tcolorbox}

\subsection{Success Rate}

A task is considered successful if and only if all three metrics achieve perfect scores:

\section{Effect of Retriever Indexing}
\label{appendix:retrieveridx}
\begin{table}[h]
\centering
\caption{Ablation study on retriever index design. T and O denote type and object, respectively. All incorporates timestamp, type, object, and website.}
\small
\begin{tabular}{lccc}
\toprule
\textbf{Scheme} & \textbf{Key} & \textbf{\textbf{$\ppref$}} & \textbf{Intent} \\
\midrule
\multirow{2}{*}{On-demand} & T + O & 0.677 & 0.542\\
  & All & 0.480 & 0.307 \\
\midrule

\multirow{2}{*}{Pre-execution} & T + O & 0.737 & 0.690\\
 & All & 0.703 & 0.677 \\
\bottomrule
\end{tabular}
\label{tab:ablation}
\end{table}
\noindent
To validate our index design choice in retriever module, we conduct an ablation study comparing alternative indexing schemes.
The original retriever extracts only \textit{type} and \textit{object} attributes from each history entry to form simplified keys.
Following common practice in information retrieval~\cite{Wu2024LongMemEvalBC, zhong2024memorybank, Maharana2024EvaluatingVL}, an alternative design incorporates all four attributes (\textit{timestamp}, \textit{type}, \textit{object}, and \textit{website}).
Results on Qwen3-80B-Instruct show that the condensed indexing scheme achieves higher preference and intent scores across both access schemes.
Adding timestamp and website attributes introduces noise rather than useful signal.
Concatenating all attributes obscures the core semantic content in the embedding, and semantic similarity-based retrievers cannot meaningfully leverage temporal or locational information.

\section{Prompt for Data Construction}
\label{appendix:dataprompt}
\begin{table*}[b]
\caption{Prompt for Domain Preference Generation}
\vspace{-5pt}
\label{tab:domainpref}
\small
\centering
\begin{tabular}{p{14cm}}
\toprule
\textbf{Prompt for Domain Preference Generation} \\
\midrule
\textbf{Travel / Accommodation / Transportation > Restaurant Reservations}\\\\
What the fields must capture\\\\
\textbf{Preference}: include 4-5 concise preferences regarding restaurant reservations\\
- MUST provide user context in a concise manner.\\
- Favorite cuisine types (e.g., Japanese, Mexican, vegetarian-friendly)\\
- Price tiers in USD or symbols (e.g., "average meal $\le$ \$40 per person")\\\\
\textbf{Patterns}: include 2–3 realistic, repeating habits such as:\\
- "Uses OpenTable when booking business dinners during weekdays"\\
- "Browses ExploreTock before special occasions"\\
\bottomrule
\end{tabular}
\end{table*}

\begin{table*}[t]
\caption{Prompt for Event Seed Generation}
\vspace{-5pt}
\label{tab:eventseed}
\small
\centering
\begin{tabular}{p{14cm}}
\toprule
\textbf{Prompt for Event Seed Generation - High Frequency} \\
\midrule

Generate realistic daily life events for month {month} of {year} based on the following user profile and requirements.\\\\

\textbf{CRITICAL INSTRUCTIONS}\\
- All events MUST align with the provided user profile\\
- The 'frequency related to' field in the user profile contains\\frequency information for each event category - strictly follow these frequencies\\
- The 'selected subdomains' field in the user profile includes\\preference information about the user's tastes and lifestyle - faithfully incorporate these preferences\\
- All events must be realistic and plausible with natural flow and timing\\
- Keep descriptions general and avoid overly specific details\\
- Avoid specific details that cannot be inferred from the profile\\(website names, brand names, etc.)\\\\

\textbf{CANCELLATION RULES}\\
- Approximately 10\% of cancellable events should be marked as cancelled\\
- Add "(cancelled)" to events that are cancelled\\
- Only cancel events where cancellation makes logical sense\\\\

\textbf{User Profile}\\
\{user profile\}\\\\

\textbf{Event Generation Guidelines}\\\\

\textbf{Core Categories}:\\\\

\textbf{Food \& Dining} \\
- Create variety: restaurant visits, cafe visits, brunch meetings, dinner with friends/colleagues \\\\

\textbf{Shopping} \\
- Online shopping ONLY - no offline shopping events\\
- Include various product categories based on user preferences\\
- Keep descriptions at appropriate level: NO more specific details\\\\

\textbf{Work \& Education} \\
- First identify user's occupation from profile: student, office worker, or freelancer\\
- This determines the base routine:\\
  - Students: school attendance, study sessions, exams\\
  - Office workers: commute\\
  - Freelancers: client work, flexible schedule, networking\\
- Include relevant special events:\\
  - Work: client meetings, conferences, presentations\\
  - Education: exams, competitions, school events, study groups\\
  - Development: online courses, workshops, certifications, skill training\\

\bottomrule
\end{tabular}
\end{table*}
\begin{table*}[t]
\caption{Prompt for Action Decomposition}
\vspace{-5pt}
\label{tab:actiondecomp}
\small
\centering
\begin{tabular}{p{14cm}}
\toprule
\textbf{Prompt for Action Decomposition} \\
\midrule

Decompose the following event into a sequence of plausible actions with timestamps. \\

Event Date: \{event\_date\} \\
Event Description: {event\_description} \\\\

\textbf{CRITICAL CONSTRAINTS:} \\
Only these 5 action types are allowed:\\
1. web search - searching for information online\\
2. web visit - visiting specific websites\\
3. purchase - buying products or services\\
4. booking - making reservations or appointments\\
5. review \& rating - writing reviews\\\\

DO NOT use any other action types.\\\\
\textbf{DECOMPOSITION GUIDELINES}:\\[4pt]
The number of actions must reflect realistic human behavior. Don't force simple tasks to have many steps, and don't oversimplify naturally complex processes. \\[4pt]

Simple Events (1-2 actions):\\
- Quick, single-purpose tasks\\
- No research or comparison needed\\\\

Complex Events (3-8 actions):\\
- Multi-stage processes requiring planning\\
- Involve research, comparison, decision-making\\
- planning trips, organizing events, major purchases\\\\

\textbf{CORE REQUIREMENTS:}\\\\

1. Temporal Logic\\\\
Actions must occur in **before/during/after** timeframes:\\
- **Before**: preparation actions (search, review reading, booking)\\
- **During**: execution actions (purchase, web visit for real-time info)\\
- **After**: follow-up actions (writing reviews, related purchases)\\[4pt]

TIMESTAMP GENERATION RULES:\\\\
- Generate timestamps around the event date\\
- Use realistic times (avoid 2-6 AM unless necessary)\\
- Format: YYYY-MM-DD HH:MM:SS\\
- Actions should be in chronological order\\[4pt]

2. Action Diversity Guidelines (DO NOT copy exactly)\\\\
- **Shopping events**: Not just `purchase'\\
\ \ - Include: product search → review reading → price comparison → purchase → review writing\\[4pt]
  
- **Travel events**: Full journey\\
\ \ - Include: destination search → booking → travel info lookup → review writing\\[4pt]

- **Dining events**: Complete experience\\
\ \ - Include: `web search' (restaurant) → `web visit' (read reviews) → `booking' (reservation) → `web visit' (location map) → `review \& rating' (write reviews) \\
\bottomrule
\end{tabular}
\end{table*}
\begin{table*}[t]
\caption{Prompt for User History Generation}
\vspace{-5pt}
\label{tab:memorybank}
\centering
\small
\begin{tabular}{p{14cm}}
\toprule
\textbf{Prompt for User History Generation} \\
\midrule

Personalize the following action into a detailed user history entry based on the user's domain profile.\\[4pt]

\textbf{Action Information:}\\
- Timestamp: \{timestamp\}\\
- Action: \{action\}\\
- Domain: \{domain\}\\
- Subdomain: \{subdomain\}\\[4pt]

\textbf{User's Domain Profile:}\\
\{domain\_profile\}\\\\[4pt]

\textbf{CRITICAL REQUIREMENTS:}\\
1. MUST reference the "preferences" field in the profile for specific brands, sizes, websites, artists, and other preferences\\
2. MUST consider the "patterns" field to understand user's routine behaviors and habits\\
3. Generate highly specific and concrete details that align with these preferences and patterns\\
4. The personalization should authentically reflect this specific user's tastes and behaviors\\\\[4pt]

\textbf{PERSONALIZATION GUIDELINES:}\\[4pt]

OUTPUT FORMAT:\\
\{\\
  "timestamp": "YYYY-MM-DD HH:MM:SS",\\
  "type": "web search, web visit, purchase, booking, review \& rating",\\
  "object": "detailed description with specifics from preferences",\\
  "website": "bare domain of the platform"\\
\}\\[4pt]

The object field MUST be extremely specific, incorporating details from the profile's preferences and patterns fields.\\
\bottomrule
\end{tabular}
\end{table*}

Tables~\ref{tab:domainpref}, \ref{tab:eventseed}, \ref{tab:actiondecomp}, and \ref{tab:memorybank} detail the prompts used in our data construction pipeline. 
These prompts guide the generation of domain preferences, daily life event seeds, action decompositions, and personalized user history entries. 
Each prompt is carefully designed to enforce realism, internal consistency, and alignment with user preferences and behavioral patterns, forming the foundation of our \textit{clarify-to-personalize} query design.

\section{Examples for History Access Schemes}
\label{appendix:schemeex}
\begin{table*}[t]
\caption{Execution log of pre-execution scheme}
\vspace{-5pt}
\label{tab:pre_example}
\small
\centering
\begin{tabular}{p{14cm}}
\toprule
\textbf{Query: Find books in my usual work category in my preferred format.} \\
\midrule
\begin{tcolorbox}
\textbf{Step 1 (History Access):} \\
\textbf{\textit{Personalization Query:}} What is the user's preferred website for finding and purchasing books? \\
\textbf{\textit{Retrieved History Index:}} 1749, 850, 854, ... \\
\textbf{\textit{Personalization Reasoning:}} Based on the histories provided, the user's preferred website for purchasing books is \textbf{Amazon}. You have made five separate book purchases, all of which were on Amazon. On November 17, 2024, you purchased 'Continuous Discovery Habits' by Teresa Torres. On October 16, ... In all instances where details are mentioned, you leveraged Amazon Prime for quick delivery.\\[4pt]

\textbf{Step 2 (History Access):} \\
\textbf{\textit{Personalization Query:}} What is the user's usual work-related book category? \\
\textbf{\textit{Retrieved History Index:}} 1893, 288 \\
\textbf{\textit{Personalization Reasoning:}} Based on the histories provided, no explicit record of work-related book categories. However, histories show interest in \textbf{product management} through online course searches on coursera.org, not through book purchases.\\[4pt]

\textbf{Step 3 (History Access):} \\
\textbf{\textit{Personalization Query:}} What is the user's preferred format for books? \\
\textbf{\textit{Retrieved History Index:}} 1593, 854, 268, ... \\
\textbf{\textit{Personalization Reasoning:}} Based on the user histories provided, the user's preferred book format is \textbf{hardcover}. All book purchases recorded in the histories are for hardcover editions. **February 18, 2024:** Purchased 'Inspired: How To Create Products Customers Love' by Marty Cagan in hardcover. **May 30, 2024:** Purchased a hardcover edition of 'INSPIRED: How to Create Products Customers Love' by Marty Cagan. **June 11, 2024:** Purchased two hardcover business books.
\end{tcolorbox}

\textbf{Step 4 (Navigation Action):} \\
\textbf{\textit{Objective:}} Enhanced task: First, go to Amazon.com. Then, search for Product Management books and filter the results to show only Hardcover editions. ... \\
\textbf{\textit{URL:}} \url{https://www.bing.com/} \\
\textbf{\textit{Reason:}} The first step of the objective is to go to Amazon.com. Since the current page is a search engine, the most direct way to navigate to Amazon.com is to type "Amazon.com" into the search bar and execute the search. \\
\textbf{\textit{Action:}} type [141] [Amazon.com] [1] \\[4pt]

\textbf{Step 5 (Navigation Action):} \\
\textbf{\textit{Objective:}} Enhanced task: First, go to Amazon.com. Then, search for Product Management books and filter the results to show only Hardcover editions. ... \\
\textbf{\textit{URL:}} \url{https://www.amazon.com/} \\
\textbf{\textit{Reason:}} The next step in the objective is to search for "Product Management" books. I will use the search bar `searchbox [8]` to input the search query "Product Management" and then proceed to click the search button to execute the search. \\
\textbf{\textit{Action:}} type [8] [Product Management] [1] \\[4pt]

\textbf{Step 6 (Navigation Action):} \\
\textbf{\textit{Objective:}} Enhanced task: First, go to Amazon.com. Then, search for Product Management books and filter the results to show only Hardcover editions. ... \\
\textbf{\textit{URL:}} \url{https://www.amazon.com/s?k=Product+Management...} \\
\textbf{\textit{Reason:}} The objective is to filter the search results to show only "Hardcover" editions. I have identified the "Hardcover" filter link (ID `16553`) within the "Book Format" section of the filtering menu. Clicking this link will apply the desired filter to the search results. \\
\textbf{\textit{Action:}} click [16215] \\[4pt]

...

\textbf{Step 9 (Navigation Action):} \\
\textbf{\textit{Objective:}} Enhanced task: First, go to Amazon.com. Then, search for Product Management books and filter the results to show only Hardcover editions. ... \\
\textbf{\textit{URL:}} \url{https://www.amazon.com/s?k=Product+Management...} \\
\textbf{\textit{Reason:}} All these steps have been successfully completed. The current page shows the filtered results as requested. Therefore, the task is finished. \\
\textbf{\textit{Action:}} stop [The search for "Product Management" books on Amazon.com has been performed, and the results have been successfully filtered to show only "Hardcover" editions. \\[4pt]

\bottomrule
\end{tabular}
\end{table*}
\begin{table*}[t]
\caption{Execution log of on-demand scheme}
\vspace{-5pt}
\label{tab:on_example}
\small
\centering
\begin{tabular}{p{14cm}}
\toprule
\textbf{Query: Find sales negotiation courses that match my typical budget and course length limits.} \\
\midrule
\begin{tcolorbox}
\textbf{Step 1 (History Access):} \\
\textbf{\textit{Personalization Query:}} user's preferred website for finding professional development courses. \\
\textbf{\textit{Retrieved History Index:}} 1113, 611, 100, ... \\
\textbf{\textit{Personalization Reasoning:}} Based on the histories provided, your preferred website for finding professional development courses is \textbf{coursera.org}. You have used coursera.org on at least nine separate occasions for various activities related to professional development, specifically focusing on sales strategy and negotiation techniques.
\end{tcolorbox}

\textbf{Step 2 (Navigation Action):} \\
\textbf{\textit{Objective:}} Query: Find sales negotiation courses that match my typical budget and course length limits for sales/negotiation. \\
\textbf{\textit{URL:}} \url{https://www.bing.com/} \\
\textbf{\textit{Reason:}} Based on the user's memory, the preferred website for finding professional development courses is \textbf{coursera.org}. Therefore, the first step is to navigate to this website. I will type coursera.org into the search bar and press enter to go to the website. \\
\textbf{\textit{Action:}} type [144] [coursera.org] [1] \\[4pt]

\textbf{Step 3 (Navigation Action):} \\
\textbf{\textit{Objective:}} Query: Find sales negotiation courses that match my typical budget and course length limits for sales/negotiation. \\
\textbf{\textit{URL:}} \url{https://www.bing.com/search?q=coursera.org...} \\
\textbf{\textit{Reason:}} I have successfully searched for the user's preferred website. The next logical step is to click on the link that leads to the Coursera homepage to begin searching for the specified courses. I will click on the link with the text "Coursera Degrees, Certificates, \& Free Online Courses" to navigate to the website. \\
\textbf{\textit{Action:}} click [15096] \\[4pt]

\textbf{Step 4 (Navigation Action):} \\
\textbf{\textit{Objective:}} Query: Find sales negotiation courses that match my typical budget and course length limits for sales/negotiation. \\
\textbf{\textit{URL:}} \url{https://www.coursera.org/?msockid...} \\
\textbf{\textit{Reason:}} The objective is to find "sales negotiation courses". The most direct way to do this is to use the search bar provided on the homepage. I will type "sales negotiation" into the search bar and press enter to see the available courses. \\
\textbf{\textit{Action:}} type [1853] [sales negotiation] [1] \\[4pt]

\begin{tcolorbox}
\textbf{Step 5 (History Access):} \\
\textbf{\textit{Personalization Query:}} user's typical budget and course length limits for sales/negotiation courses. \\
\textbf{\textit{Retrieved History Index:}} 1113, 904, 495, ... \\
\textbf{\textit{Personalization Reasoning:}} Based on the provided histories, you have looked for courses priced \textbf{under \$30 USD}. You have a limit for total course length of \textbf{under 10 hours}. You have a preferred weekly workload of \textbf{4-6 hours per week}.
\end{tcolorbox}

\textbf{Step 6 (Navigation Action):} \\
\textbf{\textit{Objective:}} Query: Find sales negotiation courses that match my typical budget and course length limits for sales/negotiation. \\
\textbf{\textit{URL:}} \url{https://www.coursera.org/search?query=sales%20negotiation...} \\
\textbf{\textit{Reason:}} I have successfully searched for "sales negotiation" courses. Now I need to filter the results to match the user's preferences. According to the user's memory, the course length should be under 10 hours. The \"Duration\" filter button [5907] is the most direct way to apply this constraint. I will click on it to see the available duration options and select the one that best fits the user's needs. \\
\textbf{\textit{Action:}} click [5907] \\[4pt]

...

\textbf{Step 14 (Navigation Action):} \\
\textbf{\textit{Objective:}} Query: Find sales negotiation courses that match my typical budget and course length limits for sales/negotiation. \\
\textbf{\textit{URL:}} \url{https://www.coursera.org/search?query=sales%20negotiation&productDuration=Less...} \\
\textbf{\textit{Reason:}} I have successfully identified multiple courses that match the user's specified criteria for website, duration, and budget. The course titled "Copilot for Beginners: Create a Sales Pitch Presentation" is particularly relevant as it explicitly mentions "Price Negotiation". Since I have found suitable options, the task is complete. I will now stop and provide the answer. \\
\textbf{\textit{Action:}} stop [The following sales negotiation courses on Coursera match your preferences (under 10 hours, under \$30): "Copilot for Beginners: Create a Sales Pitch Presentation", "Design a pitch deck with Canva", and "Create a Promotional Video using Canva".]" \\[4pt]

\bottomrule
\end{tabular}
\end{table*}

Tables~\ref{tab:pre_example} and~\ref{tab:on_example} present execution logs illustrating the two history access schemes considered in our benchmark. 
In the pre-execution scheme (Table~\ref{tab:pre_example}), the agent retrieves and reasons over multiple aspects of user history (e.g., preferred website, typical category, and format) before initiating any navigation actions. 
This enables the agent to construct a fully specified personalized objective prior to execution, after which navigation proceeds without further history access.

In contrast, the on-demand scheme (Table~\ref{tab:on_example}) interleaves history access with navigation. 
The agent initially retrieves only coarse-grained preferences (e.g., preferred platform) to start execution, and later queries user history again when additional constraints (e.g., budget limits or course duration) become necessary. 
This reactive access pattern allows the agent to refine its plan during execution but requires correctly identifying when and which personalization signals should be retrieved.

Together, these examples highlight that pre-execution emphasizes upfront personalization and planning, while on-demand prioritizes adaptive personalization during execution. 
The differing strengths and requirements of these schemes help explain why optimal history access strategies vary across agent architectures.

\section{Error Statistics}
\label{appendix:error}

\begin{figure}[h]
    \centering
    \includegraphics[width=\columnwidth]{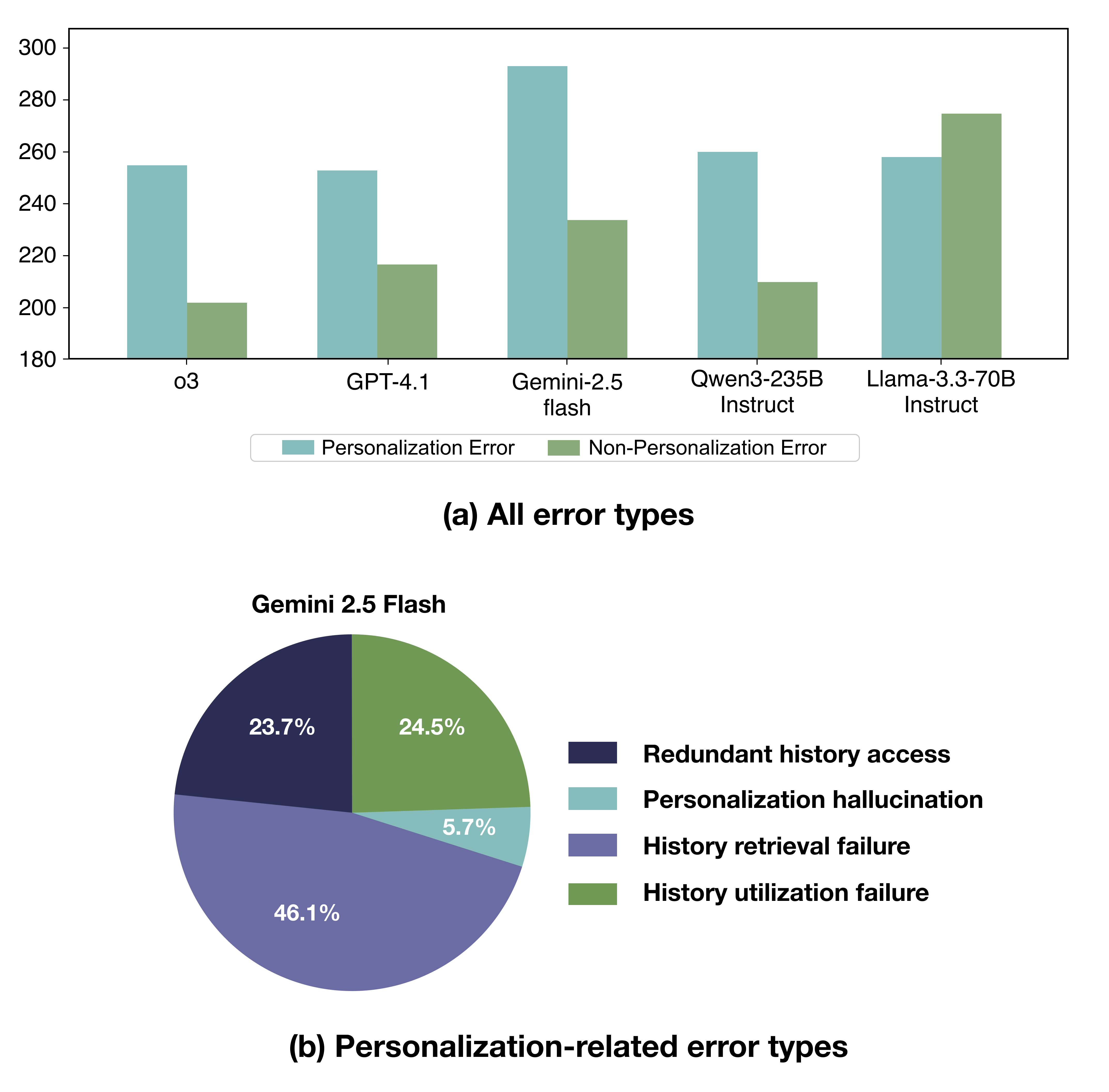}
    \caption{Error statistics for AgentOccam across backbone models (top) and a breakdown of personalization-related errors for Gemini 2.5 Flash (bottom).}
    \label{fig:error}
\end{figure}

To investigate the sources of agent failures, we conduct an error analysis based on the main results.
Errors are categorized into personalization-related errors, indicated by imperfect preference or website scores, and non-personalization-related errors, indicated by imperfect intent scores. 

Figure~\ref{fig:error}(a) reports the distribution of personalization-related and non-personalization-related errors for AgentOccam with various backbone models. 
Across all backbones, personalization-related errors constitute a substantial portion of total failures, indicating that incorrect or incomplete use of user history remains a dominant challenge even when agents successfully navigate the web. 
While the overall error counts vary by model, this pattern suggests that improvements in general web navigation alone are insufficient to address personalization failures~\cite{kim2026agenticshop}.

Figure~\ref{fig:error}(b) further breaks down personalization-related errors for Gemini~2.5~Flash. 
Among these, history retrieval failure is the most prevalent error type, accounting for nearly half of personalization errors, followed by history utilization failure and redundant history access. 
In contrast, personalization hallucination occurs less frequently. 
This breakdown highlights that personalization failures more often arise from difficulties in identifying and applying relevant historical information, such as opinions embedded in user reviews~\cite{heo2025can}, rather than fabricating nonexistent preferences.
Together, these findings reinforce the need for improved mechanisms that support effective history reasoning and utilization in personalized web agents.

\section{Case Studies of Error Types}
\label{appendix:errorex}
\begin{table*}[t]
\caption{Redundant History Access Error}
\vspace{-5pt}
\label{tab:error1}
\small
\centering
\begin{tabular}{p{14cm}}
\toprule
\textbf{Query:} Find and compare shore excursions for our upcoming cruise that match my usual preferences and budget. \\
\midrule

\textbf{Step 1 (History Access):} \\
\textit{Personalization Query:} user's preferred website for booking or browsing cruise shore excursions \\
\textit{Personalization Reasoning:} Based on the provided histories, the evidence of your preferred website for booking or browsing cruise shore excursions is your use of the \textbf{Carnival Cruises} website. ... \\[4pt]

\textbf{Step 2 (Navigation Action):} \\
\textit{URL:} \url{https://www.bing.com/} \\
\textit{Reason:} Following the user's preferences and history, the search for shore excursions should be conducted on the Carnival Cruises website. The next step is to search for the Carnival Cruises website using Bing, so I can navigate there and begin searching for shore excursions that match the user's preferences and budget. \\
\textit{Action:} type [144] [Carnival Cruises shore excursions] [1] \\[4pt]

\begin{tcolorbox}
\textbf{Step 3 (History Access):} \\
\textit{Personalization Query:} user's usual shore excursion preferences \\
\textit{Personalization Reasoning:} Based on the provided user histories, there is no direct information available regarding the user's usual shore excursion preferences. ...
\end{tcolorbox}

\begin{tcolorbox}
\textbf{Step 4 (History Access):} \\
\textit{Personalization Query:} user's cruise itinerary (dates and ports of call for the upcoming cruise) \\
\textit{Personalization Reasoning:} Based on the provided user histories, there is limited information available regarding your cruise itinerary. The only relevant detail found is from a memory dated 2023-10-21, which states: you booked a small-group, low-activity shore excursion for 2 adults on Carnival Cruises departing from Port Canaveral. ...
\end{tcolorbox}

\begin{tcolorbox}
\textbf{Step 5 (History Access):} \\
\textit{Personalization Query:} user's cruise itinerary (upcoming cruise: dates and ports of call) \\
\textit{Personalization Reasoning:} Based on the provided user histories, the available information about your upcoming cruise is as follows: Cruise Line: Carnival Cruises; Departure Port: Port Canaveral; Excursion Details: You have booked a small-group, low-activity panoramic coach sightseeing shore excursion for 2 adults. ...
\end{tcolorbox}

\textbf{Step 6 (Navigation Action):} \\
\textit{URL:} \url{https://www.bing.com/search?q=Carnival+Cruises+shore+excursions...} \\
\textit{Reason:} To fulfill the user's objective in line with their known preferences, I should proceed to the official Carnival Cruises shore excursions page. This will allow me to browse and compare available excursions, filter for low-activity or sightseeing options, and check prices, all without needing to log in or provide personal info. This step is required before any filtering or comparison based on preferences or budget. \\
\textit{Action:} click [17259] \\[4pt]

\begin{tcolorbox}
\textbf{Step 7 (History Access):} \\
\textit{Personalization Query:} user's specific shore excursion preferences (activity type, activity level, price range) \\
\textit{Personalization Reasoning:} Based on the provided user histories, several histories indicate a preference for mid-range spending (\$30–\$70 per person) and highly rated services. ...
\end{tcolorbox}

\textbf{Step 8 (History Access):} \\
\textit{Personalization Query:} user's specific shore excursion preferences (activity type, activity level, and price range) \\
\textit{Personalization Reasoning:} Based on the provided user histories, the relevant detail is that you previously searched for family-friendly activities near Lockeland Table in East Nashville, focusing on activities suitable for all ages. ... \\[4pt]

\textbf{Step 9 (Navigation Action):} \\
\textit{URL:} \url{https://www.carnival.com/shore-excursions/...} \\
\textit{Reason:} First, I must retrieve the user’s preferred website from their memory bank to support the execution of this query, and then perform the search. I should not attempt any actions that require login, phone number, or e-mail address. To match the user's known preferences (low-activity, sightseeing, mid-range budget) and efficiently compare excursions, I should apply the following filters:  
- Activity Type: Sight-seeing and Motorcoach (as panoramic coach tours are a past preference)  
- Activity Level: Easy (to ensure low-activity options)  
- Price: \$50–\$99.99 and \$100 or more (to cover mid-range and slightly higher).  
Applying these filters will narrow down excursions to those most relevant to the user’s patterns, allowing for an effective comparison. \\
\textit{Action:} click [2423] \\[4pt]

\bottomrule
\end{tabular}
\end{table*}
\begin{table*}[t]
\caption{Personalization Hallucination Error}
\vspace{-5pt}
\label{tab:error3}
\small
\centering
\begin{tabular}{p{14cm}}
\toprule
\textbf{Query:} Find monitors that match my usual monitor specs, and apply my preferred condition and review threshold; sort by price low to high. \\
\midrule

\textbf{Step 1 (History Access):} \\
\textit{Personalization Query:} user's preferred website for monitor shopping \\
\textit{Personalization Reasoning:} Based on the provided histories, the user has visited and used both eBay and Best Buy when shopping for monitors. eBay was used for price comparison, and Best Buy was checked for open-box availability. No explicit preference for either site is stated, but Best Buy appears frequently in relation to open-box deals. \\[4pt]

\textbf{Step 2 (Navigation Action):} \\
\textit{URL:} \url{https://www.bing.com/} \\
\textit{Reason:} Therefore, Best Buy is a reasonable starting point for this task. Search for “Best Buy monitors” to reach the monitor category page. \\
\textit{Action:} type [144] [Best Buy monitors] [1] \\[4pt]

\textbf{Step 3 (History Access):} \\
\textit{Personalization Query:} user's usual monitor specs (size, resolution, panel type, refresh rate etc.) \\
\textit{Personalization Reasoning:} The user typically searches for 27--32'' 4K monitors prioritizing high color accuracy, high refresh rate, and USB-C connectivity. No brand or panel-type information is available. \\[4pt]

\textbf{Step 4 (Navigation Action):} \\
\textit{URL:} \url{https://www.bing.com/search?q=Best+Buy+monitors...} \\
\textit{Reason:} Select the ``Best Buy'' link ([16375]) to navigate to Best Buy’s monitor section for applying the user’s specifications and filters. \\
\textit{Action:} click [16375] \\[4pt]

\textbf{Step 5 (History Access):} \\
\textit{Personalization Query:} user's preferred condition and minimum review threshold for monitors \\
\textit{Personalization Reasoning:} Based on the provided histories, records indicate interests in 4K monitor and purchasing monitors under open-box condition. However, there is no explicit mention of any minimum review threshold for monitor selection. \\[4pt]

\textbf{Step 6--13 (Navigation Actions):} \\
The agent sequentially selected ``United States'', navigated to ``4K Monitors'', applied the ``26.5''--31.4'' size filter'', and then the ``Open-Box'' condition filter. Finally, the results were sorted by ``Price Low to High''. These steps remain consistent with the retrieved specifications and History-grounded reasoning. \\[4pt]

\begin{tcolorbox}
\textbf{Step 14 (Navigation Action):} \\
\textit{URL:} \url{https://www.bestbuy.com/site/searchpage...} \\
\textit{Reason:} ``The objective is to find monitors that match the user’s usual specs (4K, 27--32'', open-box, high refresh rate, USB-C, high color accuracy) and apply the preferred condition and review threshold. Since the user likely wants well-reviewed products, \textbf{I should apply the ``4 Stars \& Up'' filter.}'' \\
\textit{Action:} click [4499]
\end{tcolorbox}

\textbf{Step 15 (Navigation Action):} \\
\textit{URL:} \url{https://www.bestbuy.com/site/searchpage...} \\
\textit{Reason:} The results now display open-box 4K monitors (27--32'', USB-C, well-reviewed, sorted by price). Representative examples include LG 27'' IPS UHD 4K (\$216.99, 4.8$\star$) and Dell 27'' 4K (\$239.99, 4.7$\star$). \\
\textit{Action:} stop \\[4pt]

\bottomrule
\end{tabular}
\end{table*}
\begin{table*}[t]
\caption{History Retrieval Failure Error}
\vspace{-5pt}
\label{tab:error2}
\small
\centering
\begin{tabular}{p{14cm}}
\toprule
\textbf{Query:} Check the current standings and this week's schedule for my NFL team and my MLB team. \\
\midrule

\textbf{Step 1 (History Access):} \\
\textit{Personalization Query:} What is the user's favorite or preferred NFL team? \\
\textit{Personalization Reasoning:} Based on the provided user histories, the user's favorite or preferred NFL team is the Baltimore Ravens. \\[4pt]

\begin{tcolorbox}
\textbf{Step 2 (History Access):} \\
\textit{Personalization Query:} What is the user's favorite or preferred MLB team? \\
\textit{Personalization Reasoning:} Based on the provided user histories, \textbf{there is no explicit mention of the user's favorite or preferred MLB team.}
\end{tcolorbox}

\textbf{Step 3 (History Access):} \\
\textit{Personalization Query:} What is the user's preferred website for checking sports standings and schedules? \\
\textit{Personalization Reasoning:} Based on the provided user histories, the user's preferred website for checking sports standings and schedules is Fox Sports (foxsports). If the user uses any other sports websites, this is not reflected in the current histories. \\[4pt]

\textbf{Step 4 (Navigation Action):} \\
\textit{URL:} \url{https://www.bing.com/} \\
\textit{Reason:} To proceed according to the user's preferences, I need to go to the Fox Sports website. Since there is no direct link, the most efficient step is to use the search box to search for “Fox Sports Baltimore Ravens standings and schedule” to directly reach the relevant section on Fox Sports. \\
\textit{Action:} type [141] [Fox Sports Baltimore Ravens standings and schedule] [1] \\[4pt]

\textbf{Step 5 (Navigation Action):} \\
\textit{URL:} \url{https://www.bing.com/search?q=Fox+Sports+Baltimore+Ravens+standings+and+schedule...} \\
\textit{Reason:} The user's preferred source is Fox Sports, and the top links ([17517]/[17528] and [17794]) are directly relevant to the Baltimore Ravens’ standings and schedule. To fulfill the objective, I should proceed to the Fox Sports page for the Baltimore Ravens’ schedule and standings. Link [17794] appears to be the most comprehensive for both the schedule and scores, which likely includes this week's schedule and potentially a link to standings or the standings themselves. \\
\textit{Action:} click [17794] \\[4pt]

\textbf{Step 6 (Navigation Action):} \\
\textit{URL:} \url{https://www.bing.com/search?q=Fox+Sports+Baltimore+Ravens+standings+and+schedule&form...} \\
\textit{Reason:} The objective is to retrieve both the current standings and this week's schedule for the Baltimore Ravens from Fox Sports. The most direct links to Fox Sports are [17517], [17528] (standings) and [17777] (schedule). Since the previous click on [17794] did not navigate away from Bing, the next best action is to click [17777], which is another direct Fox Sports link for the Baltimore Ravens’ schedule and may provide the required information. \\
\textit{Action:} click [17777] \\[4pt]

\textbf{Step 7 (Navigation Action):} \\
\textit{URL:} \url{https://www.foxsports.com/nfl/baltimore-ravens-team-schedule} \\
\textit{Reason:} The current page provides both the Ravens’ current standings and their schedule, including this week’s game. The record is explicitly stated at the top, and the upcoming games table includes this week's schedule. No information about an MLB team can be provided due to lack of user preference. All required information is now gathered and ready to be summarized for the user. \\
\textit{Action:} stop [Baltimore Ravens (NFL) on Fox Sports:\\
- Current standing: 1-3, 3rd in AFC North.\\
- This week’s game: vs. Houston Texans, Monday 10/5, 2:00AM at M\&T Bank Stadium, Baltimore, MD (CBS).\\
- No MLB team schedule or standings are provided, as no preferred MLB team is identified in your histories.] \\[4pt]

\bottomrule
\end{tabular}
\end{table*}
\begin{table*}[t]
\caption{History Utilization Failure Error}
\vspace{-5pt}
\label{tab:error4}
\small
\centering
\begin{tabular}{p{14cm}}
\toprule
\textbf{Query:} Check seat availability for upcoming concerts in my preferred genre with seats in my preferred location. \\
\midrule

\textbf{Step 1 (History Access):} \\
\textit{Personalization Query:} What is the user's preferred website for buying concert tickets? \\
\textit{Personalization Reasoning:} Based on the user histories provided, SeatGeek is the only website explicitly mentioned in the context of checking ticket availability. \\[4pt]

\textbf{Step 2 (History Access):} \\
\textit{Personalization Query:} What is the user's preferred music genre? \\
\textit{Personalization Reasoning:} Based on the user histories, there is no explicit mention of a preferred music genre. However, History indicates an interest in specific artists such as Kendrick Lamar and SZA, for whom the user previously searched concert venues. \\[4pt]

\textbf{Step 3 (History Access):} \\
\textit{Personalization Query:} What is the user's preferred seating location within a concert venue? \\
\textit{Personalization Reasoning:} The user's preferred seating location within a concert venue is in the lower-bowl sections near the stage. \\[4pt]

\textbf{Step 4 (History Access):} \\
\textit{Personalization Query:} What is the user's home city or the city they are interested in for finding concerts? \\
\textit{Personalization Reasoning:} Based on the user histories, the city the user is interested in for concerts is \textbf{Los Angeles}. \\[4pt]

\textbf{Step 5 (Navigation Action):} \\
\textit{URL:} \url{https://www.bing.com/} \\
\textit{Reason:} The first step of the task is to navigate to SeatGeek.com. Since the current page is a search engine, the most direct way to do this is to type “SeatGeek.com” into the search bar and press Enter. \\
\textit{Action:} type [141] [SeatGeek.com] [1] \\[4pt]

\textbf{Step 6 (Navigation Action):} \\
\textit{URL:} \url{https://www.bing.com/search?q=SeatGeek.com...} \\
\textit{Reason:} I have successfully searched for SeatGeek.com. The next step is to click the first search result, which leads to the SeatGeek homepage. \\
\textit{Action:} click [15414] \\[4pt]

\begin{tcolorbox}
\textbf{Step 7--8 (Navigation Actions):} \\
The agent searched sequentially for ``Kendrick Lamar'' and ``SZA'' concerts on SeatGeek. However, these searches were conducted at the global level \textbf{without applying the retrieved city constraint (Los Angeles)}, thereby overlooking the user's explicit location preference during reasoning and action selection.
\end{tcolorbox}

\textbf{Step 9 (Navigation Action):} \\
\textit{URL:} \url{https://seatgeek.com/sza-tickets} \\
\textit{Reason:} According to SeatGeek, there are no upcoming concerts in Los Angeles for either Kendrick Lamar or SZA. While the website mentions that SZA will be an opening act for Kendrick Lamar’s 2025 Grand National Tour, no specific dates or ticket listings are currently available. The agent stops here without verifying lower-bowl seat availability or applying the city constraint. \\
\textit{Action:} stop [No upcoming concerts found for Kendrick Lamar or SZA in Los Angeles. Seat availability could not be checked.] \\[4pt]

\bottomrule
\end{tabular}
\end{table*}

We also present representative execution logs for each of the four personalization error types identified in our analysis: Redundant History Access (Table~\ref{tab:error1}), Personalization Hallucination (Table~\ref{tab:error2}), History Retrieval Failure (Table~\ref{tab:error3}), and History Utilization Failure (Table~\ref{tab:error4}). 
These examples complement the main analysis by illustrating how each error type manifests during agent execution. 


\end{document}